\documentclass{article}
\PassOptionsToPackage{numbers}{natbib}

\usepackage[main,final]{neurips_2026}

\usepackage[utf8]{inputenc} 
\usepackage[T1]{fontenc}    
\usepackage{hyperref}       
\usepackage{url}            
\usepackage{booktabs}       
\usepackage{amsfonts}       
\usepackage{nicefrac}       
\usepackage{microtype}      
\usepackage{xcolor}         
\usepackage{graphicx}
\usepackage{subcaption}
 \usepackage{amsmath}
\title{RABBiT: Rapidly adaptive BOLD foundation model via brain-tuning for accurate zero-shot and few-shot prediction of speech-elicited responses in the brain}

\author{%
 Omer Moussa \\
  Max Planck Institute for Software Systems\\
  Saarbrücken, Germany\\
  \texttt{omoussa@mpi-sws.org} \\
  \And
  Mariya Toneva \\
  Max Planck Institute for Software Systems \\
  Saarbrücken, Germany \\
  \texttt{mtoneva@mpi-sws.org} \\
}

\begin{document}

\maketitle

\begin{abstract}

Language understanding in the brain is context-dependent, varying across experimental stimuli and individuals, which makes it difficult to build computational models that generalize across both. This calls for a foundation model of language-evoked brain activity that can capture shared structure while adapting efficiently to new participants and inputs. We introduce RABBiT (Rapidly Adaptive BOLD foundation model via BraIn-Tuning), a compact audio-to-fMRI encoder designed for accurate zero- and few-shot prediction. A comprehensive evaluation on 324 participants across multiple unseen fMRI datasets shows that RABBiT enables accurate zero-shot prediction of fMRI responses to natural speech across auditory and language-selective regions, surpassing the SOTA foundation model for fMRI and predictions based on group averages. With as little as 10 minutes of participant-specific data, RABBiT further improves performance via parameter-efficient tuning, substantially outperforming per-participant linear models. RABBiT's performance is driven by two key innovations: (1) learned region-specific attention, and (2) a decomposition of brain responses into shared and subject-specific components, combined with a brain-tuned speech backbone. In addition to supporting strong predictive accuracy, the structured, region-specific representations that RABBiT learns enable interpretability. By eliminating the need for extensive per-participant data and model fitting, RABBiT enables scalable population-level analyses of language in the human brain. \mbox{Code is available at \url{https://github.com/bridge-ai-neuro/rabbit}.} 
\end{abstract}

\medskip
\noindent\begin{minipage}{\linewidth}
  \begin{minipage}[c]{0.66\linewidth}
    \includegraphics[width=\linewidth]{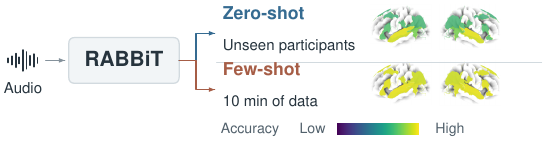}
  \end{minipage}\hfill
  \begin{minipage}[c]{0.30\linewidth}
    \captionsetup{type=figure,font=small,justification=raggedright,singlelinecheck=false,skip=0pt}
    \caption*{\textbf{Rapid transfer.} Zero-shot beats multimodal SOTA with an $\sim$80$\times$ smaller model; 10-minute adaptation beats ridge \mbox{with 1000$\times$ fewer parameters.}}
  \end{minipage}
\end{minipage}
\medskip

\begin{figure}[!t]
  \captionsetup{font=small,skip=4pt}
  \centering
  \begin{subfigure}[b]{\linewidth}
    \phantomsubcaption\label{fig:trans}
    \centering
    \includegraphics[width=0.85\linewidth]{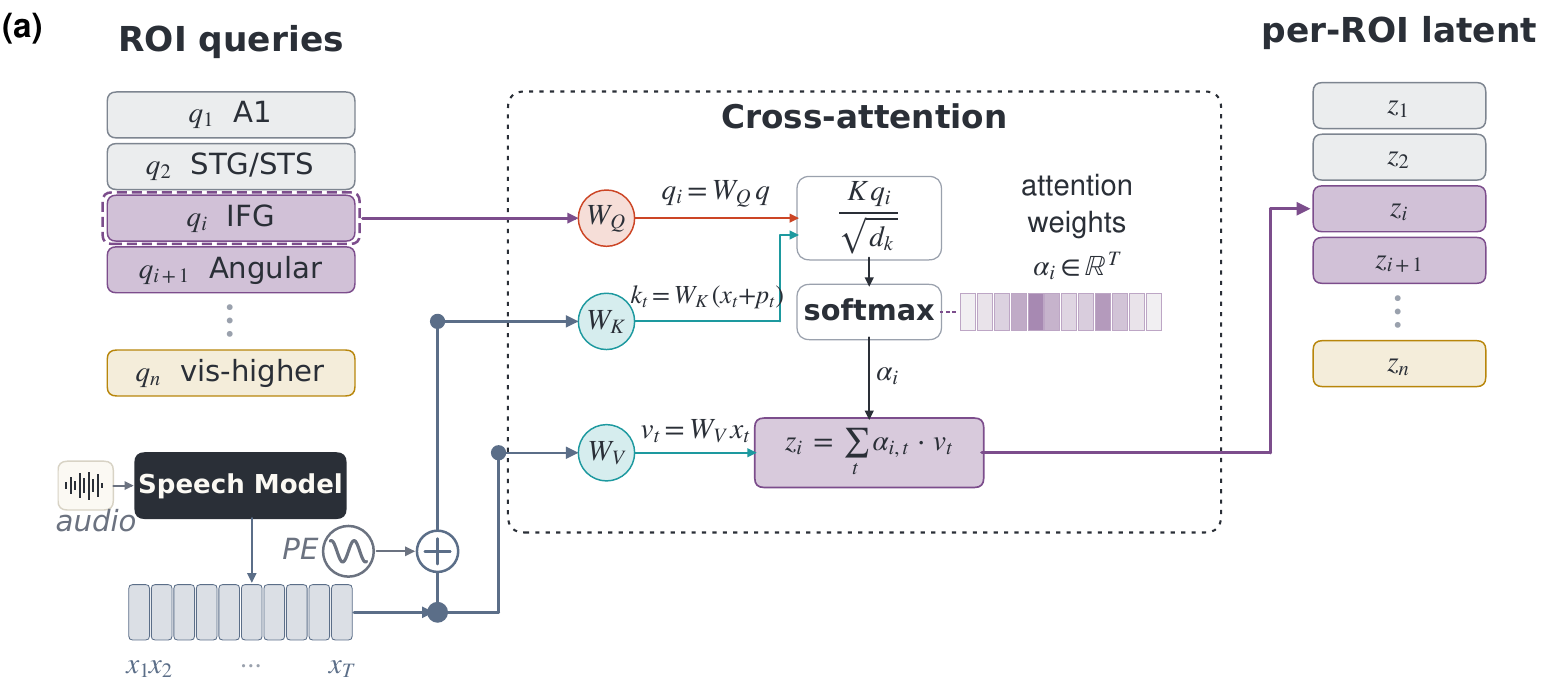}
  \end{subfigure}\\[6pt]
  \begin{subfigure}[b]{\linewidth}
    \phantomsubcaption\label{fig:training}
    \centering
    \includegraphics[width=\linewidth]{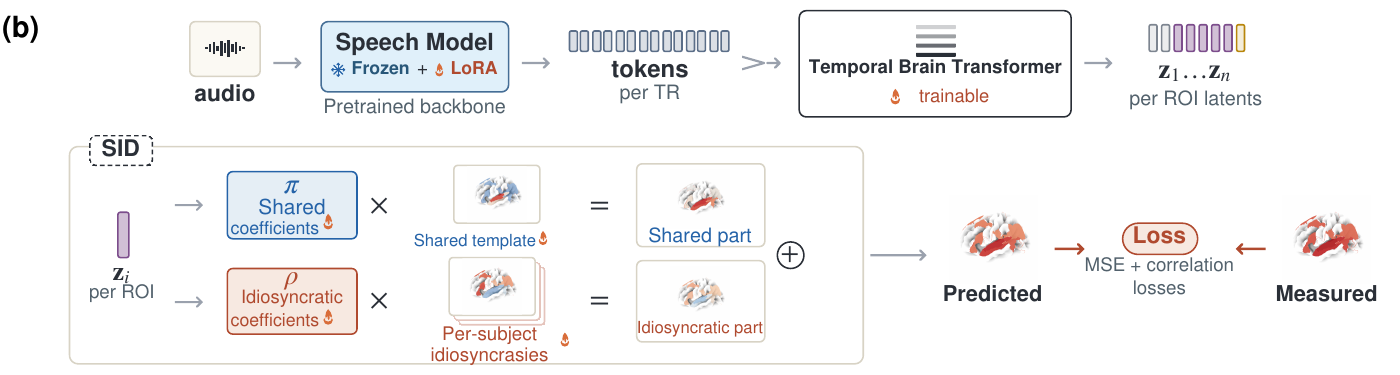}
  \end{subfigure}
  \caption{\textbf{RABBiT architecture.} (a) The cross-attention readout: learnable ROI queries attend to speech output tokens with temporally positioned keys, producing regional latents $z_i$ that feed into prediction heads. (b) Training pipeline: Audio is encoded by a speech model into a temporal token sequence; the cross-attention transformer
routes those tokens to ROI latents; the per-ROI latent passes through our SID decomposition (learnable
shared + per-subject idiosyncratic components) to form the predicted fMRI response; the loss is
applied between predicted and measured fMRI (the whole pipeline has $\sim$12.9M trainable parameters total).}
  \label{fig:overview}
\end{figure}

\section{Introduction}

\label{sec:intro}

Natural language comprehension evokes brain responses that are both reliable across listeners and strongly shaped by individual variation. Early auditory regions exhibit relatively consistent responses across participants, whereas higher-order language regions vary substantially across brains \citep{nastase2021narratives}. This creates a central challenge for language fMRI modeling: a useful encoder must capture the shared structure of language processing while remaining adaptable to individual listeners.

Most existing brain encoding approaches address only one side of this problem. Standard voxel-wise encoding models fit a separate readout for each participant \citep{huth2016natural,antonello2024scaling}. Although effective with sufficient data, these methods require collecting new fMRI recordings and fitting a high-dimensional model for every participant. More recent brain-tuning approaches improve the underlying speech representations using fMRI supervision \citep{moussa2025improving,moussa2025multibrain,vattikonda2025brainwavlm}, but still rely on participant-specific voxel-wise fitting at inference time. In parallel, recent foundation-style brain encoders improve zero-shot prediction across participants \citep{dAscoli2026TribeV2,beliy2025brainlm}, yet they are not designed for efficient adaptation to new individuals, particularly at high cortical resolution. These limitations motivate a foundation model for language-evoked brain activity: a single encoder that predicts the shared response structure of unseen listeners while supporting rapid adaptation to individual brains from limited calibration data.

We introduce \textbf{RABBiT} (\textbf{R}apidly \textbf{A}daptive \textbf{B}OLD foundation model via \textbf{B}ra\textbf{I}n-\textbf{T}uning; Fig.\ref{fig:overview}), a compact audio-to-fMRI encoder designed for accurate zero-shot predictions and efficient few-shot adaptation to new participants. RABBiT first brain-tunes a self-supervised speech backbone using LoRA adapters, adapting speech representations directly for cortical prediction. A novel \textbf{Temporal Brain Transformer (TBT)} then constructs explicit ROI-level representations by learning region-specific attention over the speech stream, replacing fixed temporal pooling and fragmented voxel-wise processing. Finally, a novel \textbf{Shared--Idiosyncratic Decomposition (SID)} maps each ROI representation to cortical activity through a low-rank shared component and a low-rank participant-specific deviation. The shared component enables accurate zero-shot population prediction, while the deviation pathway supports efficient few-shot adaptation for a new participant with only a small number of trainable parameters.


In summary, our main contributions are: \begin{itemize} \item[(i)] We introduce \textbf{RABBiT}, a compact brain-tuned audio-to-fMRI encoder that supports both zero-shot population prediction and parameter-efficient few-shot adaptation to new subjects. \item[(ii)] We propose \textbf{Temporal Brain Transformer (TBT)} as one key component of RABBiT, which learns region-specific attention over speech representations, enabling explicit ROI-level temporal routing instead of fixed temporal pooling or independent voxel-wise processing. TBT improves prediction and learns interpretable ROI queries whose geometry recovers a coarse hierarchy from auditory to higher-order language regions.
\item[(iii)] We propose \textbf{Shared--Idiosyncratic Decomposition (SID)} as a second key component of RABBiT, which learns low-rank factorizations of the neural response into shared population structure and subject-specific deviations. SID preserves zero-shot accuracy while enabling efficient adaptation to new participants from minutes of fMRI data. \item[(iv)] We evaluate on two held-out naturalistic listening cohorts totaling 324 unseen participants, and show that RABBiT's zero-shot performance exceeds that of much larger baselines, including TRIBEv2. It also reproduces canonical language-network localizers without re-training, supporting in-silico studies of speech and language.
\item[(v)] We show that few-shot SID adaptation significantly beats voxel-wise ridge encoders trained on the same calibration data. The largest gains occur in higher-order language ROIs that vary most across participants, confirming the model's central intuition. \end{itemize}

\section{Related Work}
\label{sec:relatedwork}

\textbf{Pretrained brain encoding models.} Pretrained language and speech models are widely used to predict brain responses to naturalistic language \citep{wehbe2014simultaneously,jain2018incorporating,toneva2019interpreting,schrimpf2021neural,huth2016natural,antonello2024scaling,millet2022toward,vaidya2022self,oota2024speech}. However, the dominant approach---participant-specific voxel-wise ridge regression---does not support either zero-shot prediction for unseen participants or efficient few-shot adaptation.

\textbf{Brain-tuned speech encoders.} Brain-tuning methods improve speech representations using fMRI supervision \citep{moussa2025improving,moussa2025braininter,moussa2025multibrain,vattikonda2025brainwavlm}. These approaches enhance features later used by participant-specific readouts, whereas RABBiT integrates brain-tuning as part of a full encoder designed for both zero-shot and few-shot prediction.

\textbf{Zero-shot brain foundation models.}
Recent large-scale brain encoders move toward zero-shot prediction across participants. TRIBEv2 \citep{dAscoli2026TribeV2}, for example, combines large multimodal backbones with a large transformer and subject-conditioned output heads to predict group-level responses. In contrast, RABBiT is a compact, parameter-efficient audio-to-fMRI foundation model specialized for auditory and language regions. It supports accurate zero-shot prediction while producing higher-resolution outputs (4x the resolution of TRIBEv2) and enabling efficient adaptation to new participants.

\textbf{Few-shot transfer to new participants.} Voxel-wise ridge regression remains a strong participant-specific baseline when sufficient fMRI data are available \citep{huth2016natural,antonello2024scaling}, but performs poorly in low-data regimes due to its large number of parameters. Although TRIBEv2 also supports participant adaptation, it requires fine-tuning large model components and constructing new output heads \citep{dAscoli2026TribeV2}. RABBiT instead freezes the shared encoder and adapts only a low-rank subject-specific pathway. The closest related approach, Universal Brain Encoder \citep{beliy2024wisdom}, performs few-shot adaptation for image-to-fMRI prediction, whereas RABBiT supports both speech-based zero-shot prediction and efficient few-shot adaptation within a unified framework. 

\textbf{Region-level transformer routing.} RABBiT's temporal brain transformer is inspired by transformer-based brain encoders for vision \citep{adeli2025transformer}, which use ROI queries to route visual information to cortical regions. RABBiT extends this idea to temporal speech representations, multi-subject training, and shared--idiosyncratic decoding through SID, yielding a reusable region-level temporal routing mechanism rather than a participant-specific readout.

Overall, prior work lacks a compact speech-to-fMRI encoder that simultaneously supports accurate zero-shot prediction and efficient few-shot adaptation. RABBiT addresses this gap by combining brain-tuning, ROI-level temporal routing, and a low-rank shared--idiosyncratic decomposition.

\section{Methods}
\label{sec:methods}
We motivate here the design choices for RABBiT and provide details about each model component, the training process, and the zero-shot and few-shot evaluation settings. 

\subsection{Low-rank Decomposition of Brain Responses}
\label{sec:methods-shareddev}
\subsubsection{Intuition}
\label{sec:methods-shareddev-int}

A transferable encoder for language-evoked brain activity must balance two competing objectives: capturing subject-specific structure while remaining scalable across participants. A single population-level readout is scalable but obscures reliable individual variability, whereas participant-specific readouts scale poorly with cohort size and cortical resolution \citep{dAscoli2026TribeV2}. This tradeoff is particularly acute in fMRI, where outputs are high-dimensional and subject-level data are limited.

To characterize which response components are shared across participants, we measure, for each auditory and language ROI, how well an individual's response correlates with the group average when that individual is included versus excluded from the average. Higher-order language regions show substantial degradation under leave-one-out averaging (App.\ref{app:sid_pre}), indicating stronger idiosyncratic structure, whereas early auditory regions remain comparatively stable across participants, consistent with prior inter-subject correlation analyses \citep{nastase2021narratives}.

Motivated by this structure, we introduce a novel \textbf{Shared--Idiosyncratic Decomposition (SID)} in which each ROI response is modeled as the sum of a low-rank shared component and a low-rank subject-specific deviation. The shared component captures response structure that transfers across listeners, while the deviation component models individual variability without requiring a full participant-specific voxel-wise readout. This decomposition naturally supports both transfer settings used throughout the paper. In zero-shot prediction, RABBiT combines the shared component with the average learned deviation to estimate population responses (Sec.\ref{sec:methods-zs}). In few-shot calibration, only the low-rank deviation parameters are updated, enabling efficient subject adaptation without relearning the full cortical mapping (Sec.\ref{sec:methods-fs}).

\subsubsection{Formulation}
\label{sec:methods-shareddev-form}
Shared-Idiosyncratic Decomposition (SID) decomposes each brain region-of-interest (ROI's) response into a population-shared component and a small per-subject part. Concretely, let $z_i \in \mathbb{R}^{d_o}$ be the representation used to predict the brain response (e.g., the output of temporal brain transformer; Sec.\ref{sec:methods-trans}). Then for ROI $i$ with $V_i$ voxels, SID predicts the ROI response for subject $s$ as \[ \hat{y}^{(s)}_i = \pi_i(z_i)\,  \Phi_i + \rho_i(z_i)\, \Delta_{i,s}, \] where 
$\Phi_i \in \mathbb{R}^{M \times V_i}$ is the shared basis that defines the population response subspace for ROI $i$, $\Delta_{i,s} \in \mathbb{R}^{R \times V_i}$ is the deviation basis that defines the subject-specific correction subspace, $R < M$, and the coefficient maps $\pi_i:\mathbb{R}^{d_o}\to\mathbb{R}^{M}$ and $\rho_i:\mathbb{R}^{d_o}\to\mathbb{R}^{R}$ make both terms stimulus-dependent.

SID can be viewed as a per-ROI mixed-effects readout consisting of a shared population component and a low-rank subject-specific component. The rank constraint serves as the key regularizer, enabling subject-specific flexibility only where supported by data while preventing an unconstrained voxel-wise per-subject model. Unlike independent subject-specific heads \citep{dAscoli2026TribeV2}, the deviation components form a compact basis of systematic individual differences from the population response.

We initialize the bases from the same empirical split that motivated the model: the shared bases from PCA of the mean training response, and the deviation bases from SVD of each subject's residual from that mean. This starts training from cross-subject structure in the shared term and residual subject structure in the deviation term; initialization details are given in App.\ref{app:sid-init}.

\subsection{Temporal Brain Transformer}
\label{sec:methods-trans}

SID defines the mapping from ROI-level latents $z_i$ to voxel-level brain responses, while the temporal brain transformer determines how each region constructs its latent representation from the speech stream. In contrast, voxel-wise ridge regression lacks an explicit region-level representation: it applies fixed temporal pooling (e.g., HRF convolution or lag windows \citep{huth2016natural,antonello2024scaling}) followed by independent regression for each cortical vertex. RABBiT replaces both components with learned ROI-level routing, where each region adaptively integrates temporal context and shares a common latent representation across its voxels via SID.


\textbf{Cross-attention module.} Concretely, each cross-attention block (Fig.\ref{fig:trans}) lets a learnable ROI query token $q_i$ attend over a window of speech model output tokens. The attention profile $\alpha_{i,\cdot}$ is the model's learned temporal pooling rule for ROI $i$: it specifies which parts of the speech context are used to form that region's state. Because the query is shared across subjects, this temporal profile can be inspected directly from the trained model, without fitting a separate per-subject temporal readout; full mathematical details including the choice of positional embedding in App.\ref{app:trans}.


\textbf{Temporal brain transformer.} The full temporal brain transformer stacks $L$ such cross-attention blocks over the ROI queries, following the design of \citep{adeli2025transformer}. The first layer reads backbone features into the ROI queries; later layers re-attend with updated region states, refining what each region selects from the speech stream. Multi-head attention lets a region combine complementary temporal patterns in parallel, such as faster acoustic structure and slower contextual structure, while the FFN in each block transforms the resulting ROI state. The final state is the per-ROI latent $z_i$ passed to SID. 

This architecture is powerful without becoming voxel-wise. The transformer learns temporal routing once per ROI, rather than learning separate temporal filters for every vertex and subject. Vertex-level and subject-specific structure are handled downstream by SID's low-rank shared and deviation bases. As a result, the routing module remains compact: our main model uses two cross-attention layers with eight heads per layer and hidden dimension $256$, for $\sim$2.4M parameters reused across participants; full architecture details are given in Sec.\ref{sec:methods-training}.

\subsection{Brain-tuning of the Speech Backbone}
\label{sec:methods-btune}
The temporal brain transformer learns which parts of the speech stream each ROI should use, but this routing is only as useful as the tokens it receives. A pretrained speech model provides strong acoustic and linguistic features, but its representations can be improved to predict cortical responses better with brain-tuning. Following recent work \citep{moussa2025improving,moussa2025multibrain}, we therefore brain-tune the speech backbone during RABBiT training using LoRA adapters \citep{hu2022lora}.

Brain-tuning is integrated directly into the encoder: gradients propagate from the SID readout through the temporal brain transformer into the LoRA adapters, adapting speech representations for both shared population prediction and subject-specific deviations. The adapter constraint enables parameter-efficient adaptation while preserving a shared backbone across participants, avoiding both full model retraining and reliance on fixed pretrained features.
We follow the multi-subject brain-tuning setup of \cite{moussa2025multibrain}; dataset construction and optimization details are provided in App.\ref{app:btune}.


\subsection{Dataset Details}
\label{sec:methods-datasets}
\textbf{Surface projection.} For the main model, fMRI data are projected to the FreeSurfer \texttt{fsaverage6} cortical surface (with 42K vertices per hemisphere) \citep{fischl1999high} and then parcellated to 30 Brain ROIs, totaling $\sim$41K cortical vertices. The ROIs span auditory and language cortex as well as visual, motor, and theory-of-mind control regions (App.\ref{app:roi_deets}). We also extend RABBiT to whole-cortex coverage with $\sim$75K vertices (App.\ref{app:whole_cortex}), showing that it scales very efficiently to the entire cortex.


\textbf{Training data.} We train on the \textbf{Friends} subset of CourtoisNeuroMod \citep{boyle2025friends, st2026cneuromod} --- 6 participants watching naturalistic audio-visual episodes ($\sim$565K TRs total). Additional details on participant hours, data splits, and TR duration are provided in App.\ref{app:train_data}. We also evaluated training on additional datasets, including Moth Radio \citep{lebel2023natural} ($\sim$113K TRs), but did not observe consistent performance improvements across settings (App.\ref{app:mixed_data}). This suggests that integrating heterogeneous naturalistic datasets may require more careful treatment, which we leave to future work.

\textbf{Evaluation data.} Zero-shot evaluation uses two held-out listening cohorts with no participant overlap with training: seven stories from \textbf{Narratives} \citep{nastase2021narratives} (mean length 15\,min; $\sim$30 participants per story) and the English \textbf{Le Petit Prince} audiobook \citep{Li2022LePetitPrince} (49 participants, 9 sections, $\sim$90\,min). Together, these cohorts provide 324 unseen participants for zero-shot group-response evaluation. Per-story durations, participant counts, and inclusion criteria are given in App.\ref{app:eval_data:dets}.

\subsection{Training Details}
\label{sec:methods-training}
The full training proceeds as follows. Input audio is encoded into temporal speech representations by the speech backbone (Sec.\ref{sec:methods-btune}), which are mapped to ROI-level latents by the temporal brain transformer (Sec.\ref{sec:methods-trans}). These latents are then mapped onto fMRI responses through the SID readout (Sec.\ref{sec:methods-shareddev}). Training minimizes the difference between predicted and true fMRI by jointly optimizing the SID parameters, temporal brain transformer, and LoRA-based speech adapters (Fig.\ref{fig:training}).


\textbf{Loss.} We train jointly on $S{=}6$ participants from Friends, in fsaverage6 space (Sec.\ref{sec:methods-datasets}). The loss combines a per-ROI z-scored MSE and a Pearson-correlation term. We train until convergence on held-out Friends validation clips. Loss and optimization details are in App.\ref{app:loss_train}.

\textbf{Model components details.} We use Wav2Vec2.0-base (90M params) \citep{baevski2020wav2vec} as our main speech backbone. A larger pretrained model did not yield significant gains (Sec.\ref{res:ablation}). For brain-tuning, we use LoRA rank 16 selected on Friends validation (App.\ref{app:btune}). For the brain transformer, we use 2 layers, with 8 heads and a hidden dimension of 256, also selected using the held-out Friends validation set (App.\ref{app:trans}). SID uses $M=100$ and $R=15$ after ablation (App.\ref{app:sid-dims}).



\subsection{Zero-shot Evaluation}
\label{sec:methods-zs}
For a new stimulus, RABBiT performs zero-shot inference by estimating the population-level response using the training subjects. The shared component is used directly, while the subject-specific deviation basis is replaced by its average across the $S$ training participants (see App.\ref{app:sid_pred}).

\label{sec:methods-zs_baselines}

\textbf{Linear baseline.} As a lower bound on performance, we train a linear baseline over the mean pooled outputs of the speech model (no brain transformer, no brain-tuning, and no SID).

\textbf{Per-subject Full-rank baseline.} To validate our low-rank SID approach, we also report a subject-specific heads baseline (i.e., one full readout head per-subject) while keeping everything else the same. This adds around $6$x more trainable parameters. If our model matches or beats that at zero-shot settings, this confirms our SID decomposition is sound and not an efficiency-performance trade-off.

\textbf{TRIBEv2.} As a strong baseline, we compare with the multimodal TRIBEv2 \citep{dAscoli2026TribeV2}. We use the official code to infer the model's output for the same evaluation naturalistic listening stories (which zeros-out the visual component of TRIBEv2). To represent it fairly, we compute the whole brain correlation on a subset of narratives and compare it with the reported range of the original paper (see App. \ref{app:tribe-baseline}). 

To report performance over different fMRI studies (Sec.\ref{sec:methods-datasets}), we use group-level Pearson correlation ($r_\mathrm{group}$) between the predictions and group-level mean of the study, following \citep{dAscoli2026TribeV2}. For each of these studies, we obtain and validate the audio-onset and shift (a standard practice in fMRI studies to account for the mismatch between stimulus and machine recording times); details are in App.\ref{app:eval_data:audio_onset}. 

To assess potential performance inflation from merely exploiting spatial autocorrelation in the data, we tested ROI-mean timecourses, within-ROI residuals, and unique variance in few-shot prediction. We compared our model to voxel-wise ridge, finding that the gains extend beyond predicting smooth uniform regional activity while also having unique variance above voxel-wise ridge (see App.\ref{app:spatial_controls}).

\subsection{In-silico Analyses and Ablations}
\label{sec:methods-ablation}
Beyond held-out encoding accuracy, we detail here the use of RABBiT as a model organism to localize the language network, the interpretability analyses of our temporal brain transformer, as well as the ablations used to validate the role of each component of our RABBiT model.

\textbf{In-silico language localization.} We treat the trained encoder as an in-silico subject, probing \emph{known functional selectivity} with controlled localizers used in human fMRI. We use an auditory adaptation of the functional language-localizer approach \citep{fedorenko2010new}: $18$\,s blocks of intact speech contrasted against acoustically matched, spectrally degraded (unintelligible) speech, a standard intelligibility manipulation that isolates linguistic processing above low-level acoustics \citep{scott2000identification, davisjohnsrude2003}. Each of the $32$ source-matched intact/degraded materials is presented to the model as a continuous stimulus; we slide the model's $1.49$\,s TR window with its trained hemodynamic delay basis over the block and read out the sustained (plateau) response. Per-vertex \textbf{Intact$-$Degraded} contrasts use paired $t$-tests across the $32$ source-matched pairs; we retain \texttt{fsaverage6} vertices with FDR-corrected $q<0.0001$.

\textbf{Cortical hierarchy from query embeddings.} In addition to improving accuracy and parameter-efficiency, the brain transformer enables analyses that linear voxel-wise readouts do not. The ROI query tokens are themselves data-driven learned representations of brain regions, and the relationships among them can be tested and related to literature about the brain's organization of language processing. We test this with a simple greedy walk: starting from primary auditory cortex (A1), at each step we move to the nearest unvisited ROI in query space using cosine similarity of query embeddings. If the queries have learned the right structure, the walk should retrace the cortical hierarchy of speech-to-language processing --- primary auditory $\to$ belt $\to$ STG/STS $\to$ higher-order temporal and frontal regions --- a structure that was never given to the model as supervision.
 
\textbf{Ablations.} Lastly, we ablate each design choice in turn to assess its contribution. Freezing wav2vec2.0 isolates the contribution of brain-tuning. Replacing the brain transformer with a direct linear readout isolates the importance of our brain transformer. Replacing SID with independent per-subject heads tests whether parameter sharing costs accuracy. As a robustness check, we also swap in WavLM-large \citep{chen2022wavlm} --- a larger (317M parameters) and more capable speech model --- to test whether the choice of model architecture or size is a bottleneck.

\subsection{Few-shot Evaluation}
\label{sec:methods-fs}
Given a short calibration window of paired audio-fMRI from a new participant, we perform few-shot adaptation by updating only the per-ROI deviation coefficients (the $\rho_i$ terms in Sec.\ref{sec:methods-shareddev-form}), while keeping other components frozen. This results in only $\sim$115K coefficients to update (as opposed to $\sim$200M ridge parameters). This efficiency is enabled by the SID formulation, which isolates subject-specific variability into a compact set of parameters that target idiosyncratic cortical responses not captured by the population-level component (Sec.\ref{sec:methods-shareddev}); see App.\ref{app:fs_math} for full details.

\label{sec:methods-fs_baselines}

\textbf{No SID RABBiT baseline.} To assess the role of SID in few-shot transfer, we construct a variant in which SID is replaced by linear readouts applied to the transformer outputs. We then apply the same few-shot adaptation, updating only the linear readout. This ablation tests whether structured low-rank decomposition is necessary for effective adaptation from limited calibration data (more in App.\ref{app:fs_direct}).

\textbf{Ridge baselines.} We compare RABBiT few-shot calibration against two voxel-wise ridge encoders trained on the same calibration data (details in App.\ref{app:fs_ridge}). The first uses pretrained speech features, following the canonical subject-specific encoding setup \citep{antonello2024scaling}. The second uses multi-brain-tuned features, which are expected to be more data-efficient than pretrained features \citep{moussa2025multibrain}. This second baseline tests whether RABBiT's few-shot gains come merely from inherited brain-tuned representations, or from SID's low-rank subject-adaptation pathway. We also include additional ridge baselines (ridge from an LM-tuned backbone similar to \citep{moussa2025multibrain} and ridge from the best model layer; see App.\ref{app:fs_stronger_baselines})




\section{Results}
\label{sec:results}

\label{res:setup}

Our evaluation tests transfer without the need for full participant-specific retraining. In zero-shot prediction, RABBiT uses audio input to predict group-average responses of unseen participants to held-out stories, evaluating how well the model captures shared population structure. In few-shot adaptation, a short paired audio--fMRI segment is provided for target participants and only the SID deviation pathway is updated, testing the efficiency of adaptation to subject-specific responses. We additionally perform ablations to identify the model components driving performance, and query-embedding analyses to characterize the learned cortical structure.


\begin{figure}[!tp]
  \captionsetup{font=small,skip=4pt}
  \captionsetup[subfigure]{skip=3pt}
  \centering
  \begin{subfigure}[t]{0.55\linewidth}
    \centering
    \includegraphics[width=\linewidth]{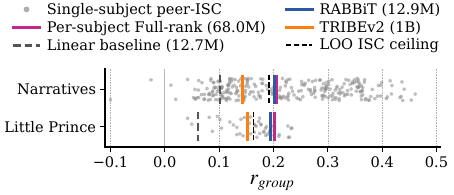}
    \caption{Zero-shot comparison across cohorts}
    \label{fig:zs_cohorts}
  \end{subfigure}\hfill
  \begin{subfigure}[t]{0.40\linewidth}
    \centering

    \includegraphics[width=\linewidth]{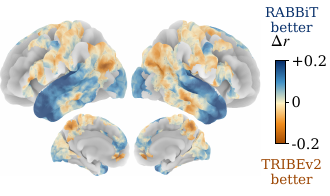}
    \caption{Zero-shot difference with TRIBEv2}
    \label{fig:brain-zs}
  \end{subfigure}\\[4pt]
  
  \begin{subfigure}[t]{0.95\linewidth}
    \centering
    \includegraphics[width=\linewidth]{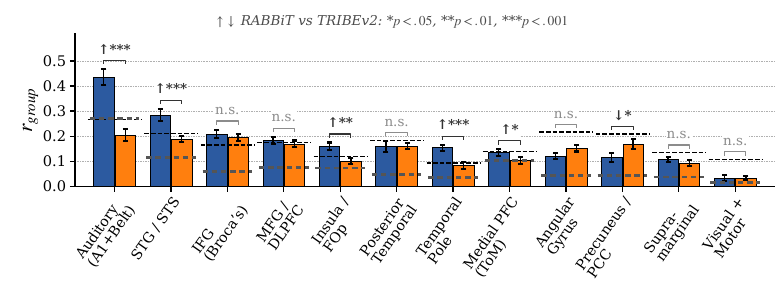}
    \caption{Per-ROI comparison}
    \label{fig:perroi-zs}
  \end{subfigure}
 \caption{\textbf{Zero-shot prediction.} (a) Group-response correlation ($r_{\mathrm{group}}$) over bilateral language ROIs on Narratives and Le Petit Prince; gray dots show single-subject peer-ISC, lines show models and leave-one-out inter-subject consistency. (b) Vertex-wise RABBiT-minus-TRIBEv2 correlation at matched resolution. (c) ROI means $\pm$ standard error across held-out stories; brackets denote paired two-sided $t$-tests against TRIBEv2. Colors and references follow (Fig.2a). RABBiT improves most in auditory and temporal regions.}
\end{figure}

\subsection{Zero-shot Evaluation}
\label{res:zs}

We evaluate RABBiT's zero-shot predictions on Narratives and {Le Petit Prince}, totaling 324 unseen participants over two cohorts (Sec.\ref{sec:methods-datasets}), and report $r_{\mathrm{group}}$ over language ROIs. Because the target is a group-average response, a relevant reference point is the leave-one-out inter-subject consistency (ISC). RABBiT surpasses this ISC estimate on both held-out cohorts (Fig.\ref{fig:zs_cohorts}). It substantially outperforms the linear baseline and the much larger TRIBEv2 model, while matching the Per-Subject head variant that uses roughly five times more parameters. The important point is therefore not only that RABBiT is accurate, but that SID preserves zero-shot accuracy while replacing independent per-subject heads with a compact shared--deviation readout.

The per-region comparison makes the pattern more diagnostic with statistical significance (two-sided t-test) measured across the different studies (Fig.\ref{fig:perroi-zs}; Fig.\ref{fig:brain-zs}). RABBiT improves most over TRIBEv2 in auditory and temporal language regions. In other language regions, such as the IFG and MFG, the two perform comparably and near the inter-subject consistency estimate. The only exception is precuneus/PCC, where TRIBEv2 performs better. Outside the speech-language network, predictions in the visual and motor regions expectedly remain near chance, due to the nature of the test stimulus.

\begin{figure}[!tp]
  \captionsetup{font=small,skip=4pt}
  \captionsetup[subfigure]{skip=3pt}
  \centering

\begin{subfigure}[t]{0.7\linewidth}
    \centering
    \includegraphics[width=\linewidth]{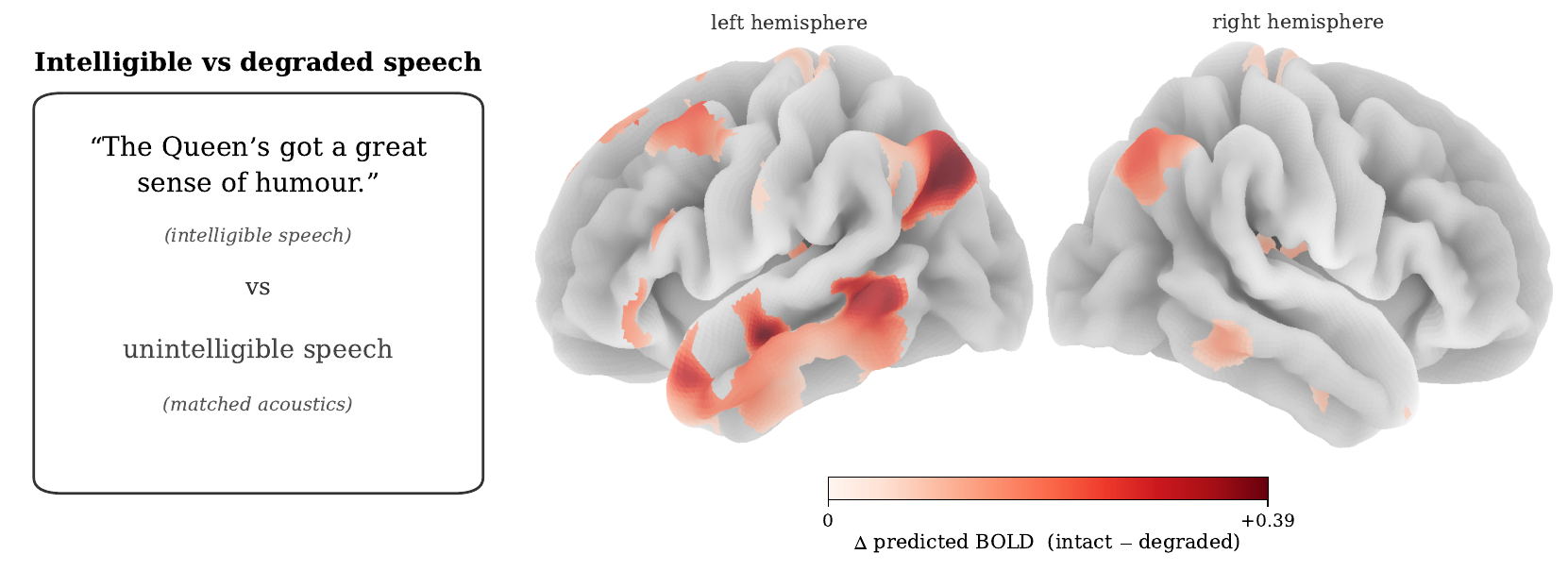}
    \caption{Language-network localization.}
    \label{fig:fed-ln}
  \end{subfigure}\\[6pt]
  \begin{subfigure}[t]{0.52\linewidth}
    \centering
    \includegraphics[width=\linewidth]{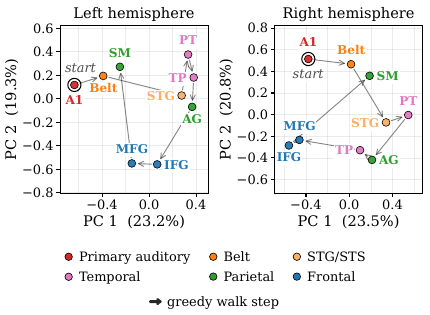}
    \caption{Learned hierarchy from query embeddings}
    \label{fig:trans_emb}
  \end{subfigure}\hfill
  \begin{subfigure}[t]{0.44\linewidth}
    \centering
    \includegraphics[width=\linewidth]{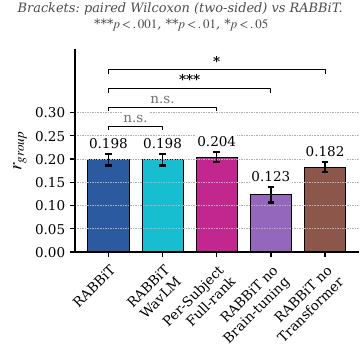}
    \caption{Ablations of model components}
    \label{fig:ablation_comp}
  \end{subfigure}

  \caption{\textbf{Functional organization and ablations.} (a) Intact$-$Degraded responses localize the canonical left-lateralized language network without localizer training. (b) Interpretation of ROI-query embeddings learned by TBT: a greedy cosine-similarity walk from A1 recovers an auditory-to-language progression. (c) Ablation of model components shows the importance of brain-tuning and TBT for RABBiT’s strong zero-shot performance; independent participant heads offer no significant gain.}
\end{figure}

\subsection{In-silico Analyses and Ablations Results}
\label{res:ablation}
\textbf{Language Localization.} RABBiT's Intact$-$Degraded contrast map reproduces a lot of the canonical, strongly left-lateralized language response \citep{scott2000identification, hickok2007cortical}. Fig.\ref{fig:fed-ln} shows the contrast map of the voxels with $q<0.0001$; intact speech is preferred in left anterior temporal cortex ($t=+5.7$), angular gyrus ($+6.2$), middle frontal gyrus ($+5.8$) and superior temporal gyrus/sulcus ($+2.2$). After naturalistic brain-tuning of its (acoustic in nature) backbone, RABBiT recovers the language network's spatial signature without training on controlled localizers or non-naturalistic data.

\textbf{Learned Query Embeddings.} The learned ROI queries provide a direct view into the temporal brain transformer. Because each query specifies the speech context an ROI attends to, their embedding geometry reflects the model's cortical organization. Starting from A1 and greedily stepping to the nearest unvisited ROI in query space (cosine similarity), we recover a coherent trajectory without any hierarchy or spatial supervision. The resulting path progresses from primary and belt auditory cortex through STG/STS and temporal association areas toward parietal and frontal language regions (Fig.\ref{fig:trans_emb}). Permuting the query-to-ROI assignment disrupts this hierarchy (App.\ref{app:query_permutation}).


\textbf{Component Ablations.} We further perform ablations to quantify component contributions. Removing brain-tuning yields the largest drop in zero-shot performance (to $\sim60\%$; Fig.\ref{fig:ablation_comp}, $p<0.001$), and removing the temporal brain transformer also significantly degrades performance ($p<0.05$). In contrast, replacing SID with larger per-subject heads provides no significant improvement despite $\sim$5$\times$ more parameters, and swapping Wav2Vec2.0 for WavLM-large has negligible effect. Together, these results indicate that brain-tuning and ROI-level routing are essential for zero-shot prediction, while SID provides compact subject modeling without sacrificing performance.


\subsection{Few-shot Evaluation}
\label{res:fs}

We further test whether RABBiT can improve over zero-shot prediction for a new participant using limited fMRI calibration data. For each participant, we freeze the speech backbone, temporal brain transformer, and shared SID pathway, and update only the low-rank per-ROI deviation parameters ($\sim$115K parameters; Sec.\ref{sec:methods-fs}). We compare against voxel-wise ridge regression trained on the same calibration data using either pretrained or brain-tuned speech features \citep{huth2016natural,antonello2024scaling,moussa2025multibrain}. This is a strong baseline: ridge fits a full voxel-wise subject-specific readout, whereas RABBiT only learns a compact correction to its zero-shot prediction. We evaluate primarily on long narrative segments ($\sim$50 min) to allow for increasing amounts of calibration data (see App.\ref{app:fs_calib}). 

\begin{figure}[!t]
  \captionsetup{font=small}
  \centering
  \includegraphics{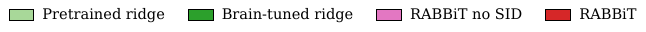}\\[4pt]
  \begin{subfigure}[b]{0.53\linewidth}
    \phantomsubcaption\label{fig:fewshot}
    \centering
    \includegraphics[width=\linewidth]{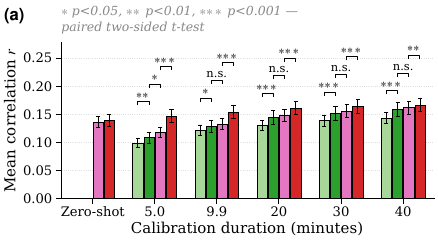}
  \end{subfigure}\hfill
  \begin{subfigure}[b]{0.38\linewidth}
    \phantomsubcaption\label{fig:brain-fs}
    \centering
    \includegraphics[width=\linewidth]{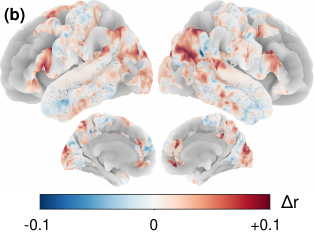}
  \end{subfigure}
  \captionsetup{skip=4pt}
  \caption{\textbf{Few-shot adaptation.} (a) Correlation (mean $\pm$ standard error across participants) on \textit{21styear} with 5--40 minutes of calibration. RABBiT calibrates SID deviations; baselines fit full readouts to the same data. RABBiT few-shot performance improves over zero-shot with only 5 minutes of calibration data and outperforms both ridge baselines (b) Vertex-wise correlation gain over zero-shot after 10 minutes, concentrated in higher-order language regions. App.\ref{app:fs_add_results} reports per-ROI comparisons and similar gains on another dataset.}
\end{figure}

RABBiT few-shot improves over zero-shot with only 5 minutes of calibration and outperforms both ridge baselines at every tested duration from 5 to 40 minutes (Fig.\ref{fig:fewshot}). Similar trends hold on an additional dataset (15 min; App.\ref{app:fs_add_results}). Among the ridge baselines in Fig.\ref{fig:fewshot}, brain-tuned features perform best; this baseline improves with more data but starts below zero-shot and requires substantially more calibration to match performance. In contrast, RABBiT begins from a strong population model and uses calibration data exclusively to refine subject-specific deviations, yielding roughly three orders of magnitude fewer trainable parameters than ridge while achieving better prediction. This low-calibration advantage also holds against LM-tuned ridge and pretrained layer-8 ridge across all auditory and language regions (App.\ref{app:fs_stronger_baselines}).

Few-shot gains concentrate in the idiosyncratic higher-order language regions identified in Sec.\ref{sec:methods-shareddev} (Fig.\ref{fig:brain-fs}); per-ROI comparisons and additional calibration results are provided in App.\ref{app:fs_add_results}.

\section{Discussion and Conclusion} \label{sec:discussion} 

The central insight of RABBiT is that accurate personalization in predicting language-evoked fMRI does not require fitting a new high-dimensional model for every participant. Instead, RABBiT separates prediction into a shared speech-to-brain mapping and a compact subject-specific correction pathway. This structure naturally supports both transfer regimes explored in this work: zero-shot prediction through the shared component, and few-shot adaptation through low-rank subject deviations.

RABBiT bridges a gap between prior brain-tuning approaches and recent foundation-style brain encoders. Brain-tuning improves speech representations for fMRI prediction, but still relies on participant-specific voxel-wise readouts \citep{moussa2025improving,moussa2025multibrain,vattikonda2025brainwavlm}. Foundation-style models move toward population-level prediction, but often depend on large backbones, subject-specific heads, or coarse spatial outputs \citep{dAscoli2026TribeV2,beliy2025brainlm}. In contrast, RABBiT shows that a compact audio-only encoder can achieve highly accurate group-level prediction while supporting efficient subject adaptation from only minutes of calibration data. Importantly, calibration updates only the compact SID deviation pathway rather than relearning the full speech-to-brain mapping.

The regional structure of these gains is particularly informative. Higher-order language regions both exhibit the largest inter-subject variability and benefit most from few-shot adaptation, indicating that individual differences are structured rather than noise-like. SID captures this structure in a compact representation that can be efficiently adapted with limited data.

The temporal brain transformer provides a complementary level of structure by learning ROI-specific temporal queries over the speech stream. Beyond improving prediction, these learned queries recover a coarse progression from auditory to higher-order language regions without explicit supervision, suggesting a promising framework for probing cortical language hierarchies. Finally, we also show that RABBiT can reproduce some canonical neuroscience findings without being explicitly trained for that, indicating a strong potential for such models to be used for initial scientific hypothesis testing.

Together, these results position RABBiT as a foundation model of language-evoked brain activity. The model predicts responses for unseen listeners without additional fMRI, adapts efficiently to new individuals, and exposes interpretable regional structure. More broadly, our findings suggest that effective brain foundation models should explicitly separate shared and individual variability and model how different brain regions integrate information over time.

\paragraph{Limitations and Future Work.} RABBiT is evaluated primarily on English naturalistic listening and auditory/language regions. Extending the evaluation to multilingual, conversational, reading, and multimodal settings will assess the generality of its shared and subject-specific representations. Second, the training cohort remains relatively small, and preliminary experiments combining datasets did not yield proportional gains, suggesting that scaling training across datasets may require improved handling of inter-dataset variability. Third, exploiting spatial autocorrelation in fMRI remains a limitation of such models even though our controls reveal predictive gains beyond ROI-wide means (App.\ref{app:spatial_controls}); future work will include explicit training objectives and designs to mitigate it further. Finally, we only partially explored scaling to larger speech backbones; a more systematic investigation of model scaling remains a direction for future work.

\clearpage
\renewcommand{\acksection}{\section*{Acknowledgements}}
\begin{ack}
Funded by the European Union (ERC, BrainAlign, 101221906). Views and opinions expressed are however those of the author(s) only and do not necessarily reflect those of the European Union or the European Research Council. Neither the European Union nor the granting authority can be \mbox{held responsible for them.}

\medskip
{\centering
\includegraphics[width=0.3\linewidth]{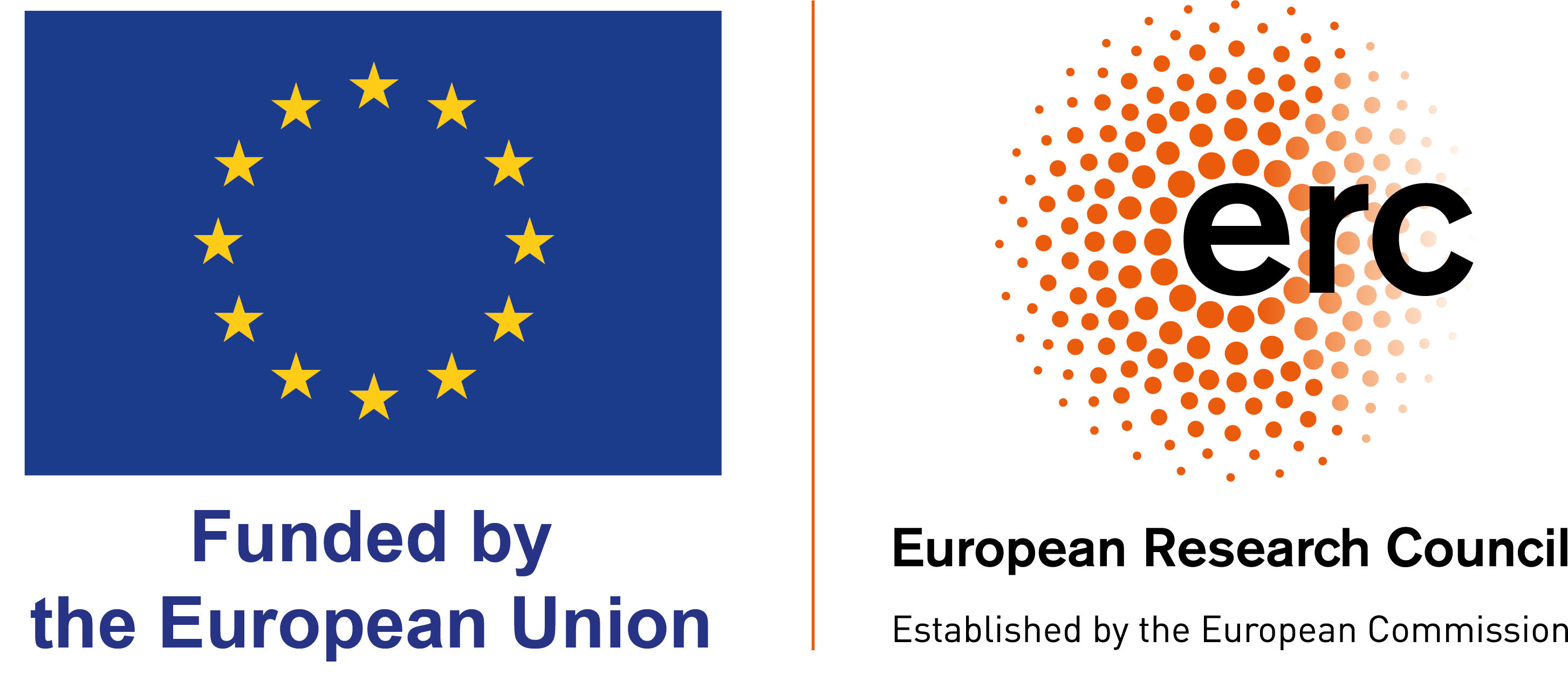}\par
}
\end{ack}

\newpage
\bibliographystyle{plain}

\bibliography{main}




\appendix

\section{Datasets and preprocessing}
\label{app:data}

All experiments use a common cortical space so that the trained encoder can be evaluated easily across datasets and participants without refitting. This section details the  parcellation of ROIs, the additional details of training and evaluation data, and the exploratory mixed-data run that motivated training on Friends alone (App.\ref{app:mixed_data}).

\subsection{Brain ROIs Details}
\label{app:roi_deets}

For the main model, fMRI is projected to FreeSurfer's \texttt{fsaverage6} surface ($40{,}962$ vertices per hemisphere; mean inter-vertex spacing $\sim$2\,mm) \citep{fischl1999high, fischl2012freesurfer} and restricted to $30$ bilateral HCP-MMP1 ROIs \citep{glasser2016multi}, totaling $\sim$$41{,}394$ vertices ($\sim$half of the full \texttt{fsaverage6} cortex). The $30$ ROIs cover the canonical auditory and language network plus sensorimotor, default-mode/theory-of-mind, and visual regions. We expand beyond language regions to test our model ability and efficiency to learn around 50\% of the cortical surface while being parameter-efficient. It stress tests the brain transformer's ability to scale the output size with substantial increase in the trainable parameters as with the full linear ridge models.

The $30$ ROIs are organized into $15$ functional groups (each with LH and RH instances), with HCP-MMP1 parcels enumerated for the auditory and language groups:
\begin{itemize}
\setlength\itemsep{0pt}
  \item \textbf{Auditory.} \emph{aud\_primary} (A1); \emph{aud\_belt} (A4, A5, LBelt, MBelt, PBelt).
  \item \textbf{Superior temporal and temporal pole.} \emph{stg\_sts} (STGa, STSda, STSdp, STSva, STSvp, STV); \emph{temporal\_pole} (TE1a, TE1m, TE1p, TE2a, TE2p, TGd, TGv); \emph{posterior\_temp} (PHT, TPOJ1, TPOJ2, TPOJ3).
  \item \textbf{Parietal language.} \emph{angular\_gyrus} (PGi, PGs, PGp, PFm); \emph{supramarginal} (PFop, PFt, PFcm, PSL).
  \item \textbf{Frontal language.} \emph{ifg} (44, 45, 47l); \emph{mfg\_dlpfc} (55b, IFJa, IFSp, SFL, 8Av, 8BL).
  \item \textbf{Default-mode and theory-of-mind controls.} \emph{mpfc\_tom}; \emph{precuneus\_pcc}.
  \item \textbf{Sensorimotor and insular controls.} \emph{motor}; \emph{insula\_fop}.
  \item \textbf{Visual controls.} \emph{visual\_early} (V1--V4); \emph{visual\_higher} (V6, V6A, V7, V8, LO1, LO2, MT, MST, FST parcels).
\end{itemize}

\subsubsection{Whole-cortex extension and prefrontal coverage}
\label{app:whole_cortex}
The main mask already includes prefrontal regions relevant to language and theory of mind: IFG/Broca, MFG/DLPFC, and mPFC. To test whether the architecture can accommodate broader coverage, we train a whole-cortex variant with one query per ROI, predicting $\sim$75K cortical vertices with $17.3$M trainable parameters. The original model predicts $\sim$41K vertices with $12.9$M trainable parameters. Extending coverage preserves the same region-query and SID formulation.

Table\ref{tab:whole_cortex_frontal} reports zero-shot correlations averaged over four Narratives stories (\textit{21styear}, \textit{slumlordreach}, \textit{tunnel}, and \textit{sherlock}). Within the original mask, both models are evaluated on the same vertices. Frontal correlations remain of similar magnitude, and the whole-cortex model also predicts activity across broader lateral PFC, cingulate/ACC, orbitofrontal, and frontal-pole regions, extending beyond the original mask. Its correlation exceeds the leave-one-out inter-subject consistency (ISC) estimate in each listed region. ISC provides a reliability reference, rather than a strict ceiling for prediction of a group-average response (App.\ref{app:zs_acc}).

\begin{center}
\begin{minipage}{\linewidth}
\captionsetup{type=table}
\centering
\small
\begin{tabular}{lrrr}
\toprule
Frontal region & RABBiT & Whole-cortex & ISC \\
\midrule
\multicolumn{4}{l}{\textit{Within the original mask}} \\
IFG/Broca (44/45/47l) & $+0.1883$ & $+0.1743$ & 0.0995 \\
MFG/DLPFC & $+0.1826$ & $+0.1669$ & 0.1153 \\
mPFC (ToM) & $+0.1243$ & $+0.1385$ & 0.0703 \\
\midrule
\multicolumn{4}{l}{\textit{Additional frontal coverage}} \\
Lateral PFC (8/9/46) & --- & $+0.1165$ & 0.0828 \\
Cingulate/ACC & --- & $+0.0943$ & 0.0432 \\
Orbitofrontal cortex & --- & $+0.0774$ & 0.0326 \\
Frontal pole (BA10) & --- & $+0.0741$ & 0.0385 \\
\bottomrule
\end{tabular}
\caption{\textbf{Frontal prediction with expanded cortical coverage.} Mean vertex-wise correlation across four Narratives stories. The whole-cortex extension covers $\sim$75K vertices with $17.3$M trainable parameters. Dashes indicate broader regions not fully covered by the original output mask. ISC is computed on the corresponding vertices.}
\label{tab:whole_cortex_frontal}
\end{minipage}
\end{center}

\subsection{Training data}
\label{app:train_data}
We train on the Friends subset of CourtoisNeuroMod \citep{boyle2025friends, st2026cneuromod}: $S{=}6$ participants watching naturalistic audio-visual TV episodes at TR\,=\,$1.49$\,s, totaling $\sim$$565{,}000$ TRs ($\sim$$234$ hours, $\sim$$39$ hours per participant). RABBiT is audio-only, so only the audio track is used as input; we use the audio of an audio-visual corpus to keep input modality and recording pipeline comparable to our held-out audio-only evaluation cohorts.

The fMRI is provided in volumetric MNI space at 2\,mm resolution and projected onto \texttt{fsaverage6} surface vertices, with the $\sim$$41{,}394$ vertices belonging to the $30$ ROIs retained per clip. The released CourtoisNeuroMod data are already z-scored per voxel per run, so no further temporal centering is applied --- consistent with the SID shared-basis initialization (App.\ref{app:sid-init}). Within each subject's clips we hold out four fixed clips for validation and four separate fixed clips for testing (held-out evaluations on training participants); the split is identical across subjects so validation losses are directly comparable. Transformer width and depth, LoRA rank, and SID anchor strength are selected using Friends validation data. Narratives and Le Petit Prince are reserved for transfer evaluation and are not used to choose these hyperparameters.

\subsection{Evaluation data}
\label{app:eval_data}

The two held-out cohorts together include $324$ unseen participant--story instances --- $275$ across the $7$ Narratives stories \citep{nastase2021narratives} (the column total of Table\ref{tab:narratives_per_story}, with each story drawn from an independent participant pool) plus $49$ unique Le Petit Prince participants \citep{Li2022LePetitPrince} who heard all $9$ audiobook sections. None overlap with the Friends training cohort.

\subsubsection{Per-cohort details}
\label{app:eval_data:dets}

\textbf{Narratives \citep{nastase2021narratives}.} We use $7$ stories from the Narratives corpus (OpenNeuro \texttt{ds002345}), each with an independent participant pool: four longer stories and three shorter, large-cohort stories (Table\ref{tab:narratives_per_story}). 

\begin{table}[h]
\centering
\small
\begin{tabular}{lrrrl}
\toprule
Story & TRs & Duration (min) & Participants & Calibrated audio onset (s) \\
\midrule
\textit{21styear}      & 2249 & 56.2 & 20 & 0.0 \\
\textit{slumlordreach} & 1210 & 30.0 & 18 & 3.0 \\
\textit{tunnel}        & 1037 & 25.5 & 23 & 0.0 \\
\textit{sherlock}      &  725 & 18.0 & 36 & 3.0 \\
\textit{pieman}        &  300 &  7.5 & 82 & 0.0 \\
\textit{notthefall}    &  400 &  9.7 & 56 & 1.5 \\
\textit{prettymouth}   &  475 & 11.9 & 40 & 1.5 \\
\midrule
\textbf{total / mean}  & \textbf{6396} & \textbf{$\sim$159} & \textbf{275} & --- \\
\bottomrule
\end{tabular}
\caption{\textbf{Narratives held-out stories used in zero-shot evaluation.} TRs at the corpus's native TR\,=\,$1.5$\,s. Participant counts are independent across stories; the column total is the participant--story instance count. Calibrated audio onsets are validated as in App.\ref{app:eval_data:audio_onset}.}
\label{tab:narratives_per_story}
\end{table}

\textbf{Le Petit Prince \citep{Li2022LePetitPrince}.} The English audiobook portion of OpenNeuro \texttt{ds003643}: $49$ native English speakers listening to $9$ scanning sections totaling $\sim$$93$ minutes at TR\,=\,$2.0$\,s on a 3T GE Discovery MR750 (a different scanner from CourtoisNeuroMod or Narratives), providing an out-of-distribution test of RABBiT's transfer beyond the Narratives recording conventions. The released preprocessed runs already strip the leading silence at audio onset, so the audio plays at scanner-time zero and no per-section onset calibration is required (the A1-LH shift sweep peaks at $0$ or $+1$ TR for $6$ of the $9$ sections; the $3$ noise-dominated sections are evaluated at the same $0$-TR convention). We use an audio onset of $0$ for the LPP dataset.

\subsubsection{Audio--fMRI alignment and onset validation}
\label{app:eval_data:audio_onset}
Audio--fMRI alignment must account for stimulus-file timing and the delay between speech and the BOLD response. We use reliable stimulus-onset metadata directly when available. Otherwise, we estimate alignment from a sample predicted A1 response profile and the measured A1 response. Timing parameters are fixed per story or dataset and applied identically across its participants, rather than selected independently for each participant or ROI.

We distinguish the audio-file onset timestamps in Table\ref{tab:narratives_per_story}, reported in seconds, from prediction-to-fMRI alignment shifts, reported in TRs in Table\ref{tab:alignment_offsets}. The latter align model predictions to measured BOLD, including hemodynamic timing. Thus, a $0$-second file onset for \textit{21styear} is compatible with a $3$-TR prediction-alignment shift; these are different timing quantities.

\paragraph{Offset consistency across model variants.} Table\ref{tab:alignment_offsets} reports prediction-alignment shifts estimated from A1 response profiles for four studies. RABBiT and its ablated variants yield identical estimates for each study, so the comparison between these variants does not involve different selected shifts.

\begin{center}
\begin{minipage}{\linewidth}
\captionsetup{type=table}
\centering
\small
\begin{tabular}{lrrrr}
\toprule
Model & \textit{21styear} & \textit{slumlordreach} & \textit{tunnel} & LPP \\
\midrule
RABBiT & 3 & 2 & 0 & 0 \\
Per-subject full-rank & 3 & 2 & 0 & 0 \\
RABBiT no brain-tuning & 3 & 2 & 0 & 0 \\
RABBiT no transformer & 3 & 2 & 0 & 0 \\
\bottomrule
\end{tabular}
\caption{\textbf{Alignment-offset consistency across RABBiT variants.} Each study has identical prediction-alignment shifts (TRs) across variants; Table~\ref{tab:narratives_per_story} lists audio-file onsets.}
\label{tab:alignment_offsets}
\end{minipage}
\end{center}

\begin{samepage}
\paragraph{Sensitivity to a one-TR perturbation.} To test whether the RABBiT--TRIBEv2 comparison depends on precise alignment, we perturb the reported offset by $-1$ and $+1$ TR and measure the mean zero-shot correlation gap across language regions and participants. Table\ref{tab:offset_sensitivity} reports this check for one Narratives story and one LPP section. The gap remains positive under both perturbations in both examples, indicating that the observed advantage is not specific to the exact reported offset.

\begin{center}
\begin{minipage}{\linewidth}
\captionsetup{type=table}
\centering
\small
\begin{tabular}{lrrr}
\toprule
Evaluation segment & Gap at $-1$ TR & Gap at reported offset & Gap at $+1$ TR \\
\midrule
Narratives, Story 1 & $+0.0952$ & $+0.1002$ & $+0.0966$ \\
LPP, Section 1 & $+0.1145$ & $+0.1238$ & $+0.1376$ \\
\bottomrule
\end{tabular}
\caption{\textbf{Sensitivity of the zero-shot model comparison to alignment.} The gap is RABBiT minus TRIBEv2 mean prediction correlation across language regions and participants; positive values favor RABBiT. The advantage persists at both $\pm1$-TR perturbations.}
\label{tab:offset_sensitivity}
\end{minipage}
\end{center}
\end{samepage}

\subsection{Training with mixed data}
\label{app:mixed_data}
We added the Moth Radio Hour subset of \citep{lebel2023natural} (we chose $6$ participants, $\sim$$113{,}000$ TRs at TR\,=\,$2.0$\,s, $\sim$$63$ hours) to the Friends mix, yielding a $12$-participant combined cohort with the same SID architecture (the shared bases $\Phi$ was obtained from both dataset mixtures while the Idiosyncratic part remains per subject.). During training, we forced a $1{:}2$ Moth-to-Friends batch ratio per training block to account for the size mismatch between the datasets. Across the four narrative story zero-shot evaluation, the combined model achieves a story-mean vertex-wise $r$ of $0.108$ versus $0.117$ for Friends-only (an $\sim$$8\%$ relative drop). The combined model matches Friends-only on Friends-side validation loss, ruling out undertraining. We have also tried per-subject Full-rank alternatives (see Sec.\ref{sec:methods-zs}) and got similar behavior. This suggests that mixing datasets during training is not straightforward and scaling law doesn't apply linearly regardless of the training data sources, complementary to the findings in \citep{moussa2025multibrain}; thus, we defer this to future work.

\section{Shared–Idiosyncratic diagnostic and SID details}
\label{app:sid_deets}





The SID readout is motivated by a simple empirical observation: some cortical regions are well predicted by the group-mean response across listeners, while others lose reliability sharply when the target listener is excluded from that mean.

\subsection{The shared--idiosyncratic diagnostic}
\label{app:sid_pre}

For each ROI $i$ and subject $s$ with stimulus-aligned response $y^{(s)}_i \in \mathbb{R}^{T \times V_i}$, we compute the vertex-averaged Pearson correlation between $y^{(s)}_i$ and the cohort-mean response $\bar{y}_i$ (\emph{including} subject $s$), then repeat with the leave-one-out cohort-mean $\bar{y}^{(-s)}_i$. The drop between the two measures how much of subject $s$'s response is captured only when they are part of the mean: a large drop means the LOO mean predicts subject $s$ poorly, so the response in ROI $i$ is more idiosyncratic than shared.

\begin{figure}[h]
    \centering
    \begin{subfigure}[b]{0.9\textwidth}
        \includegraphics[width=1\textwidth]{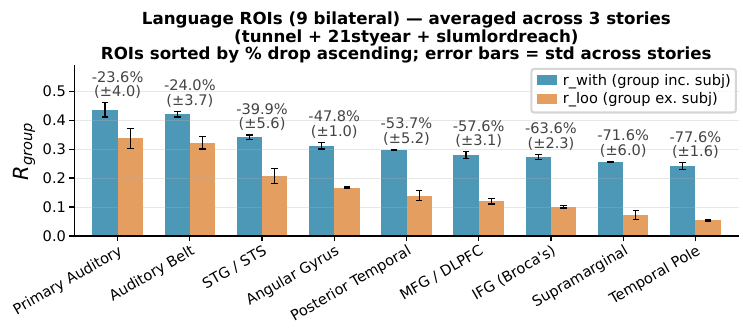}
    \end{subfigure}
    \caption{\textbf{Shared--idiosyncratic diagnostic across language and auditory ROIs.} Vertex-averaged correlation of each subject's response with the cohort mean when included versus excluded, averaged across participants and three Narratives stories (\textit{21styear}, \textit{tunnel}, \textit{slumlordreach}). Auditory and STG/STS responses remain stable; higher-order language responses decline substantially.}
    \label{fig:sid_diagnostic}
\end{figure}

Fig.\ref{fig:sid_diagnostic} plots this contrast across the auditory and language ROIs, averaged over three Narratives stories. Early auditory regions (A1, auditory belt) and superior temporal cortex remain stable under leave-one-out averaging --- their responses are largely shared, consistent with high inter-subject correlations in naturalistic listening \citep{nastase2021narratives}. Higher-order language and theory-of-mind regions (IFG, MFG/DLPFC, supramarginal, angular gyrus, mPFC) show substantial drops, indicating subject-specific structure that a population mean alone cannot capture. SID's per-subject deviation pathway gives these idiosyncratic regions a bounded budget of subject-specific structure to draw on; the shared component handles the regions on the shared end.

\subsection{SID Bases construction and initialization}
\label{app:sid-init}

The SID bases are warm-started from an empirical decomposition of the training fMRI. For each ROI $i$ and training subject $s$, we run a per-ROI PCA on $20{,}000$ subsampled TRs from that subject's clips, taking the top $M=100$ principal components as a per-subject basis $\Phi^{(s)}_i$. The shared basis $\Phi_i$ is the cross-subject average re-orthogonalized via SVD: $\Phi_i = \mathrm{SVD}\big(S^{-1}\sum_s \Phi^{(s)}_i\big)$. The idiosyncratic (deviation) basis $\Delta_{i,s}$ is the SVD of $\Phi^{(s)}_i - \Phi_i$ truncated to the top $R=15$ right singular vectors --- the dominant directions in which subject $s$ departs from the shared basis. During joint training the bases are anchored to their initialization with $\mathcal{L}_{\mathrm{anchor}} = \lambda_{\Phi} \sum_i \|\Phi_i - \Phi_i^{(0)}\|_F^2 + \lambda_{\Delta} \sum_{i,s} \|\Delta_{i,s} - \Delta_{i,s}^{(0)}\|_F^2$.

\paragraph{Anchor-weight selection.} We tie $\lambda_{\Phi}=\lambda_{\Delta}=\lambda_{\mathrm{anchor}}$ rather than tuning the two weights independently. We train candidates on a subset of the Friends training data, sweep a logarithmic grid, and select the weight with the lowest held-out validation loss. Table\ref{tab:anchor_validation} reports the best validation loss and the final regularization-term magnitude for each candidate. The selected $\lambda_{\mathrm{anchor}}=10^{-3}$ balances limiting basis drift with allowing the bases to adapt; stronger and weaker penalties both yield higher validation loss.

\begin{center}
\begin{minipage}{\linewidth}
\captionsetup{type=table}
\centering
\small
\begin{tabular}{rrr}
\toprule
$\lambda_{\mathrm{anchor}}$ & Best validation loss & Final regularization term \\
\midrule
$10^{-5}$ & 1.88091 & 0.000000 \\
$10^{-4}$ & 1.87686 & 0.000003 \\
$\mathbf{10^{-3}}$ & \textbf{1.87621} & 0.000027 \\
$10^{-2}$ & 1.87818 & 0.000313 \\
$10^{-1}$ & 1.88159 & 0.002030 \\
$1$       & 1.88419 & 0.008586 \\
\bottomrule
\end{tabular}
\caption{\textbf{SID anchor selection on Friends validation.} The shared and deviation anchor weights are tied. Bold marks the selected weight and minimum validation loss. Final regularization contributions are reported to six decimal places.}
\label{tab:anchor_validation}
\end{minipage}
\end{center}


\textbf{ More details on the Idiosyncratic (Deviation) bases $\Delta_{i,s}$.} The per-subject deviation basis is built from each subject's residual in basis-coefficient space: we form $\Phi^{(s)}_i - \Phi_i$, take its SVD, and keep the top $R=15$ right singular vectors as $\Delta_{i,s} \in \mathbb{R}^{R \times V_i}$. By construction, $\Delta_{i,s}$ spans the dominant directions in which subject $s$'s within-ROI response geometry departs from the cohort-shared basis.


\subsection{Choice of shared and deviation ranks}
\label{app:sid-dims}
The shared rank $M$ is chosen as a coverage threshold: at $M=100$ the cumulative explained variance of the per-subject PCA reaches $0.993$ averaged across $6$ subjects $\times 30$ ROIs (Table\ref{tab:sid_var_K}); only bilateral motor cortex ($V \approx 5{,}000$) is below $95\%$. Auditory cortex saturates by $M{=}20$, language ROIs by $M{=}50$. $M{=}100$ is conservative; $M{=}50$ would lose only $\sim$$2\%$ mean coverage.

\begin{table}[h]
\centering
\small
\begin{tabular}{rrrrrr}
\toprule
$M$ & mean cum.\ var. & std & min ROI & median ROI & max ROI \\
\midrule
 5  & 0.737 & 0.114 & 0.463 & 0.755 & 0.978 \\
10  & 0.836 & 0.095 & 0.587 & 0.863 & 0.997 \\
20  & 0.914 & 0.071 & 0.697 & 0.938 & 1.000 \\
30  & 0.947 & 0.055 & 0.763 & 0.970 & 1.000 \\
50  & 0.975 & 0.036 & 0.843 & 0.990 & 1.000 \\
75  & 0.988 & 0.022 & 0.900 & 0.997 & 1.000 \\
\textbf{100} & \textbf{0.993} & \textbf{0.015} & \textbf{0.933} & \textbf{0.999} & \textbf{1.000} \\
\bottomrule
\end{tabular}
\caption{\textbf{Cumulative explained variance of the per-subject shared bases as a function of $M$.} $n{=}180$ cells per row ($6$ subjects $\times 30$ ROIs). The default $M{=}100$ retains $99.3\%$ of within-ROI variance on average.}
\label{tab:sid_var_K}
\end{table}

The deviation rank $R=15$ is a regularization budget, not a coverage threshold: at $R{=}15$ the deviation basis retains only $29\%$ of the per-subject residual variation. To reach $80$--$90\%$ residual energy would require $R \approx 50$--$75$, granting the readout disproportionate per-subject flexibility. A joint $M{=}200$, $R{=}30$ ablation at the same backbone and transformer mostly probes the $R$ axis ($M{>}100$ adds at most $0.7\%$ coverage, while $R{:}15{\to}30$ nearly doubles deviation flexibility; in our experiments, this scaling does not improve held-out performance.


\section{RABBiT Architecture and Training Details }
\label{app:training}
This section gives the full architecture, loss, and optimization details that reproduce the headline configuration.

\subsection{Speech backbone and brain-tuning}
\label{app:btune}

The speech backbone supplies RABBiT with a temporal stream of frame-level audio tokens that the brain transformer routes into ROI latents. We use Wav2Vec2.0-base \citep{baevski2020wav2vec} as the backbone in all reported runs --- a $90$M-parameter self-supervised model trained on $960$ hours of LibriSpeech --- because it is the standard acoustic SSL backbone in the brain-tuning lineage \citep{moussa2025improving, moussa2025multibrain, vattikonda2025brainwavlm} and produces a feature sequence at $50$\,Hz ($20$\,ms per frame), giving the brain transformer a fine-grained acoustic stream to pool over.

\textbf{LoRA adapters.} We brain-tune the backbone with parameter-efficient LoRA \citep{hu2022lora} of rank $r=16$, selected using the held-out Friends validation set (Table\ref{tab:lora_rank}). Adapters are inserted only in the linear projections of the audio-encoder transformer blocks. The pretrained weights of every module remain frozen; only the LoRA matrices are updated in the backbone. This keeps the backbone shared across all training subjects, with approximately $1.2$M trainable adapter parameters and a $90$M frozen body.

\textbf{Learning rate and Freeze warmup.} We use a learning rate of $1e-4$ with a linear warmup period of $10\%$ and then a linear decay over epochs. For the first $3$ epochs the backbone is held entirely frozen and only the brain transformer and SID readout are updated; LoRA training begins thereafter, governed by the same early-stopping criterion (patience $5$).

\textbf{Training hardware} We use AdamW optimizer to update the trainable paramters of the model; we train with a batch size of 256 for 30 epochs with early stopping available with a 5 epoch patience. The training takes around 10 hours for 30 epochs on a single H200 GPU, with early stopping usually stopping the training at 10-13 epochs.

\paragraph{LoRA rank selection.} The rank sweep uses models trained on $40\%$ of the training data. Rank $16$ yields the highest Friends validation correlation (Table\ref{tab:lora_rank}). We also report LPP all-vertex correlations as a transfer check. Although LPP peaks at rank $8$, we retain rank $16$ selected on Friends validation.

\begin{center}
\begin{minipage}{\linewidth}
\captionsetup{type=table}
\centering
\small
\begin{tabular}{lrr}
\toprule
LoRA rank & Friends validation $r$ & LPP all-vertex $r$ \\
\midrule
4 & 0.2236 & 0.0646 \\
8 & 0.2330 & 0.0704 \\
\textbf{16 (selected)} & \textbf{0.2385} & 0.0615 \\
32 & 0.2314 & 0.0618 \\
\bottomrule
\end{tabular}
\caption{\textbf{LoRA rank selected on Friends validation.} All candidates use $40\%$ of the training data. Bold marks the configuration chosen by validation correlation; the LPP column is a transfer check and does not enter selection.}
\label{tab:lora_rank}
\end{minipage}
\end{center}

\subsection{Temporal brain transformer architecture}
\label{app:trans}

The temporal brain transformer is a stack of transformer cross-attention blocks (Fig.\ref{fig:trans} in main text). We chose $L=2$ layers for it and a hidden dimension $d_o=256$, $H=8$ heads, and FFN dimension $4 d_o = 1024$ with GELU non-linearity. Backbone features at $50$\,Hz are projected to $d_o$ via a learned linear layer with sinusoidal positional encoding added to keys at each cross-attention. The $30$ ROI queries ($q_i \in \mathbb{R}^{d_o}$, one per bilateral ROI; $\sim$$7.7$K params) are initialized to zero and combined with a learned per-ROI positional embedding. Each block applies self-attention over the queries, then cross-attention from queries to backbone tokens (values gathered without positional info), then FFN. The second-layer self-attention is the load-bearing one for cross-ROI coordination: its inputs carry stimulus-specific information from the first cross-attention, which lets each query condition its second-pass extraction on the global brain state.

\paragraph{Width and depth selection.} We select the hidden dimension and depth by maximizing correlation on the held-out Friends validation set (Tables\ref{tab:trans_hidden}--\ref{tab:trans_depth}). A width of $256$ and depth of $2$ attain the highest validation correlation among the tested values. The accompanying LPP all-vertex evaluations confirm that these choices transfer to an unseen cohort; they are not used to choose dataset-specific optima.

\begin{center}
\begin{minipage}{\linewidth}
\captionsetup{type=table}
\centering
\small
\begin{tabular}{lrr}
\toprule
Hidden dimension $d_o$ & Friends validation $r$ & LPP all-vertex $r$ \\
\midrule
64 & 0.2419 & 0.0884 \\
128 & 0.2407 & 0.0939 \\
\textbf{256 (selected)} & \textbf{0.2459} & 0.1102 \\
512 & 0.2404 & 0.0834 \\
\bottomrule
\end{tabular}
\caption{\textbf{Transformer width selected on Friends validation.} Bold marks the selected hidden dimension and highest validation correlation. LPP values are reported only to assess transfer.}
\label{tab:trans_hidden}
\end{minipage}
\end{center}

\begin{center}
\begin{minipage}{\linewidth}
\captionsetup{type=table}
\centering
\small
\begin{tabular}{lrr}
\toprule
Depth $L$ & Friends validation $r$ & LPP all-vertex $r$ \\
\midrule
\textbf{2 (selected)} & \textbf{0.2459} & 0.1102 \\
4 & 0.2446 & 0.1017 \\
8 & 0.2437 & 0.0990 \\
10 & 0.2435 & 0.1023 \\
\bottomrule
\end{tabular}
\caption{\textbf{Transformer depth selected on Friends validation.} Bold marks the selected depth and highest validation correlation. Increasing depth does not improve validation performance or LPP transfer in this sweep.}
\label{tab:trans_depth}
\end{minipage}
\end{center}

\subsubsection{ROI-query assignment controls}
\label{app:query_permutation}
The greedy walk in Fig.\ref{fig:trans_emb} operates on learned query embeddings. To test whether the structure shown there is meaningful, we therefore test whether its anatomical ordering depends on the learned assignment of embeddings to ROI labels. 

\paragraph{Exhaustive permutation of the labeled hierarchy.} Within each hemisphere, we enumerate all $9!=362{,}880$ assignments of the nine learned embeddings to language-network ROI labels. For every assignment, we repeat the same greedy cosine-similarity walk, starting at the A1 label. We score anatomical ordering by the Spearman correlation between walk position and a fixed processing-order ranking: A1, belt, STG/STS, posterior temporal cortex, temporal pole, angular gyrus, supramarginal gyrus, IFG, and MFG/DLPFC. The learned assignments attain ordering scores of $0.950$ (LH) and $0.917$ (RH). Only $0.11\%$ and $0.36\%$ of permutations, respectively, achieve scores at least as high (Table\ref{tab:query_hierarchy_null}). These percentages measure equally strong anatomical ordering, not exact reproduction of the reference path.

\begin{center}
\begin{minipage}{\linewidth}
\captionsetup{type=table}
\centering
\small
\begin{tabular}{lrr}
\toprule
Hierarchy-order statistic & Left hemisphere & Right hemisphere \\
\midrule
Learned assignment, Spearman $\rho$ & 0.950 & 0.917 \\
Permutations with $\rho\geq\rho_{\mathrm{learned}}$ & $0.11\%$ & $0.36\%$ \\
\bottomrule
\end{tabular}
\caption{\textbf{Exact permutation control for query-hierarchy ordering.} Each hemisphere uses all $362{,}880$ embedding-to-label assignments, with the greedy walk always starting at A1.}
\label{tab:query_hierarchy_null}
\end{minipage}
\end{center}

\paragraph{Displacement from the learned path.} We also compare each permuted path $P$ with the learned reference path $P_{\mathrm{ref}}$ using an ROI-identity-preserving Earth Mover's Distance (EMD), treating the paths in the paper as the reference and computing the distance to it for all path permutations:
\[
D(P,P_{\mathrm{ref}})=\frac{1}{9}\sum_{i=1}^{9}
\left|\operatorname{pos}_{P}(i)-\operatorname{pos}_{P_{\mathrm{ref}}}(i)\right|.
\]
Here $\operatorname{pos}_{P}(i)$ is the position of ROI $i$ in path $P$. Preserving ROI identity makes $D$ the mean positional displacement in walk steps. The learned path has distance $0$ by construction; a fully reversed reference path gives $4.44$ steps. Across the exhaustive permutation controls, the mean displacement is $2.33$ steps in each hemisphere (Table\ref{tab:query_path_displacement}).

\begin{center}
\begin{minipage}{\linewidth}
\captionsetup{type=table}
\centering
\small
\begin{tabular}{lrr}
\toprule
Hemisphere & Learned path & Mean under permutation \\
\midrule
Left & 0.00 & 2.33 \\
Right & 0.00 & 2.33 \\
\bottomrule
\end{tabular}
\caption{\textbf{Query permutation displaces the labeled hierarchy.} Mean ROI-position displacement from the learned path (Fig.\ref{fig:trans_emb}), in walk steps, across all $9!$ embedding assignments per hemisphere.}
\label{tab:query_path_displacement}
\end{minipage}
\end{center}

\paragraph{Query-only intervention during prediction.} We additionally evaluate six randomly chosen permutations of the $30$ ROI-query embeddings during zero-shot prediction on three Narratives stories. The speech backbone, all other transformer weights, and SID readouts remain fixed; no model is retrained. We average permutation scores per ROI and compare them with the learned assignment using paired, two-sided Wilcoxon signed-rank tests across the $30$ ROIs, with Benjamini--Hochberg (BH) correction. Between $25$ and $26$ ROIs show lower mean prediction correlation in each story (Table\ref{tab:query_predictive}). Together with the geometric controls, this intervention shows that query reassignment disrupts the labeled hierarchy and reduces predictive accuracy.

\begin{center}
\begin{minipage}{\linewidth}
\captionsetup{type=table}
\centering
\small
\begin{tabular}{lrrc}
\toprule
Narratives story & ROIs with lower correlation & $q$ (BH) & \\
\midrule
\textit{21styear} & $26/30$ & $1.06\times10^{-6}$ & $***$ \\
\textit{slumlordreach} & $26/30$ & $6.15\times10^{-7}$ & $***$ \\
\textit{sherlock} & $25/30$ & $5.86\times10^{-5}$ & $***$ \\
\bottomrule
\end{tabular}
\caption{\textbf{Zero-shot effects of query-only permutation.} Counts indicate ROIs with lower mean correlation over six random reassignments. The $q$-values are BH-corrected paired Wilcoxon tests across the $30$ ROIs; $***$ denotes $q<0.001$.}
\label{tab:query_predictive}
\end{minipage}
\end{center}

\subsection{SID prediction and zero-shot inference}
\label{app:sid_pred}

After the brain transformer produces a per-ROI latent $z_i \in \mathbb{R}^{d_o}$, the SID readout maps $z_i$ to vertex-level fMRI in two parallel pathways: shared coefficients $\pi_i(z_i) \in \mathbb{R}^M$ combine with the shared basis $\Phi_i$, and deviation coefficients $\rho_i(z_i) \in \mathbb{R}^R$ combine with the per-subject deviation basis $\Delta_{i,s}$ (both $\pi_i, \rho_i$ are linear). For ROI $i$ and subject $s$,
\[
\hat{y}^{(s)}_i \;=\; \pi_i(z_i)\, \Phi_i \;+\; \rho_i(z_i)\, \Delta_{i,s}.
\]

For a held-out subject the idiosyncratic (deviation) basis $\Delta_{i,s}$ is unknown; we replace it with the cohort average $\bar{\Delta}_i = \frac{1}{S}\sum_s \Delta_{i,s}$, giving $\hat{y}_i = \pi_i(z_i)\,\Phi_i + \rho_i(z_i)\,\bar{\Delta}_i$. This single substitution is the only change between training-time and zero-shot inference. Replacing the cohort-averaged deviation pathway with zero (predicting only the shared component) recovers approximately half the signal of the averaged-deviation prediction across the auditory and language ROIs of the held-out cohorts: even when no individual training-subject's deviation basis matches the new participant, the cohort average captures population-level structure the new participant inherits as a population draw.

\subsection{Loss and training details}
\label{app:loss_train}
Training optimizes the full encoder end-to-end against measured fMRI responses across the six Friends training participants (App.\ref{app:train_data}), using the loss
\[
\mathcal{L} \;=\; \sum_i \big(\,s_i\,\mathcal{L}^{\mathrm{L2}}_i \;+\;\lambda_{\mathrm{corr}}\,\mathcal{L}^{\mathrm{corr}}_i\,\big) \;+\; \mathcal{L}_{\mathrm{anchor}}
\]
where $\mathcal{L}^{\mathrm{L2}}_i$ is the per-ROI vertex-wise squared error between predicted and measured z-scored fMRI, $\mathcal{L}^{\mathrm{corr}}_i$ is the per-ROI correlation loss ($1$ minus the per-ROI vertex-mean Pearson correlation), and $s_i$ is a per-ROI learned scalar that lets each ROI calibrate its own squared-error weight independently. The anchor term ($\mathcal{L}_{\mathrm{anchor}}$) is defined in App.\ref{app:sid-init}.

\textbf{Why the per-ROI scaled-correlation form.} The squared-error term alone produces predictions well-calibrated in magnitude but only weakly correlated with the target on the higher-order ROIs whose response geometry is harder to recover; the correlation term alone produces predictions that are correlated but unscaled, and trains poorly because correlation gradients are scale-invariant. The scaled-correlation form combines the two, so each ROI gets gradient pressure on both correlation and magnitude, with a per-ROI learned scale that absorbs cross-ROI variation in noise floor (auditory ROIs have a much higher signal-to-noise ratio than e.g. mPFC and need less L2 weight to be well-fit). We use $\lambda_{\mathrm{corr}} = 1$ throughout. Our choice for this value was also done empirically, similar to the anchor search (see App.\ref{app:sid-init})

\textbf{Optimization.} Parameters are optimized with AdamW. The brain transformer, coefficient maps, and LoRA adapters share a base learning rate; the SID bases use a smaller learning rate that keeps them close to their PCA/SVD initialization while still allowing fine-grained refinement under gradients from the brain-tuned backbone.



\section{Zero-shot evaluation and Baselines}
\label{app:zs_baselines}

Zero-shot evaluation tests whether the trained encoder predicts the reliable group response of unseen listeners without using any fMRI from those listeners.

\subsection{Accuracy metrics and the inter-subject consistency}
\label{app:zs_acc}
We report two metrics per (model, ROI, story). The group-level correlation $r_{\mathrm{group}}^{(i)} = \mathrm{vertex\_mean}_{v \in i}\big[\mathrm{corr}_t(\hat{y}[t,v], \bar{y}^{(\text{cohort})}[t,v])\big]$ targets the cohort-mean fMRI; the per-subject consistency $r^{(i,s)}$ targets each individual subject's fMRI and is averaged across the $N$ subjects in the cohort. Both inputs are z-scored per vertex over time before correlation. The leave-one-out inter-subject correlation (LOO ISC), computed by holding out each subject in turn and correlating the mean fMRI of the other $N{-}1$ with the held-out subject, is the cleanest stimulus-only predictor the cohort itself provides and serves as the natural consistency (ceiling) for $r^{(i,s)}$. It is not a strict ceiling for $r_{\mathrm{group}}$: a model whose prediction is identical for every subject can exceed LOO ISC in $r_{\mathrm{group}}$ when the cohort is small or the ROI is highly stereotyped, because the cohort-mean target is itself noisy at small $N$ but the model's prediction is noiseless across subjects (RABBiT's A1 reaches $r_{\mathrm{group}} \approx 0.55$ vs LOO ISC $\approx 0.27$ on the headline Narratives stories).

All models are evaluated on a calibrated timebase that accounts for stimulus onset and hemodynamic timing (App.\ref{app:eval_data:audio_onset}). Speech at time $t$ elicits a delayed BOLD response, typically $4$--$6$ seconds later \citep{antonello2024scaling}; TRIBEv2 incorporates hemodynamic compensation in its training and prediction pipeline. The story-specific prediction-alignment shifts used in the RABBiT control analyses are listed in Table\ref{tab:alignment_offsets}.

\subsection{Linear and Per-Subject Full-Rank baselines}
\label{app:indp_baselines}

\textbf{Linear baseline.} The Linear baseline removes everything that distinguishes RABBiT from a frozen-feature linear probe: a frozen Wav2Vec2.0-base, mean-pooling over time within each TR-aligned window to a single $768$-dim vector, and a per-ROI linear layer mapping that vector to all $V_i$ vertices in ROI $i$. There is no temporal brain transformer and no SID; the only trainable parameters are the per-ROI linear readouts. Any gap between RABBiT and the Linear baseline must come from brain-tuning, the temporal brain transformer, or SID, not from the backbone or the vertex parcellation.

\textbf{Per-Subject Full-Rank baseline.} The Per-Subject Full-Rank baseline preserves the brain-tuned backbone and the temporal brain transformer but replaces the SID readout with one independent full-rank linear layer per ROI per training subject (no shared--deviation factorization). This adds $\sim$$5\times$ more total parameters than RABBiT (per-subject heads grow linearly with the cohort) but otherwise matches RABBiT exactly. If RABBiT matches or beats this variant in zero-shot, the SID compression has no measurable accuracy cost --- the bounded shared-plus-deviation parametrization is at least as expressive as independent per-subject heads at this cohort size.

\subsection{TRIBEv2 baseline}
\label{app:tribe-baseline}

TRIBEv2 \citep{dAscoli2026TribeV2} is a tri-modal encoder over video, audio, and text trained on multiple large public fMRI corpora, with frozen LLaMA / Wav2Vec-BERT / V-JEPA backbones, learned MLP projectors per modality, an $8$-layer cross-modal transformer, and subject-conditioned linear output heads predicting onto $\sim$$20{,}000$ fsaverage5 vertices at $1$\,Hz with a $5$\,s hemodynamic offset baked into its training contract. \emph{Vision modality is zeroed at our inference and must be disclosed:} our held-out cohorts are pure naturalistic listening, so TRIBEv2's vision input is replaced with a zero tensor and the visual contribution to the cross-modal combiner reduces to a constant baseline. (TRIBEv2 was trained with $30\%$ modality dropout, so missing-modality input is in-distribution, but the model still optimizes for the joint case.) RABBiT outperforming TRIBEv2 on visual ROIs is therefore an audio-only-vs-audio-only comparison.

\begin{figure}[h]
    \centering
    \begin{subfigure}[b]{0.75\textwidth}
        \includegraphics[width=1\textwidth]{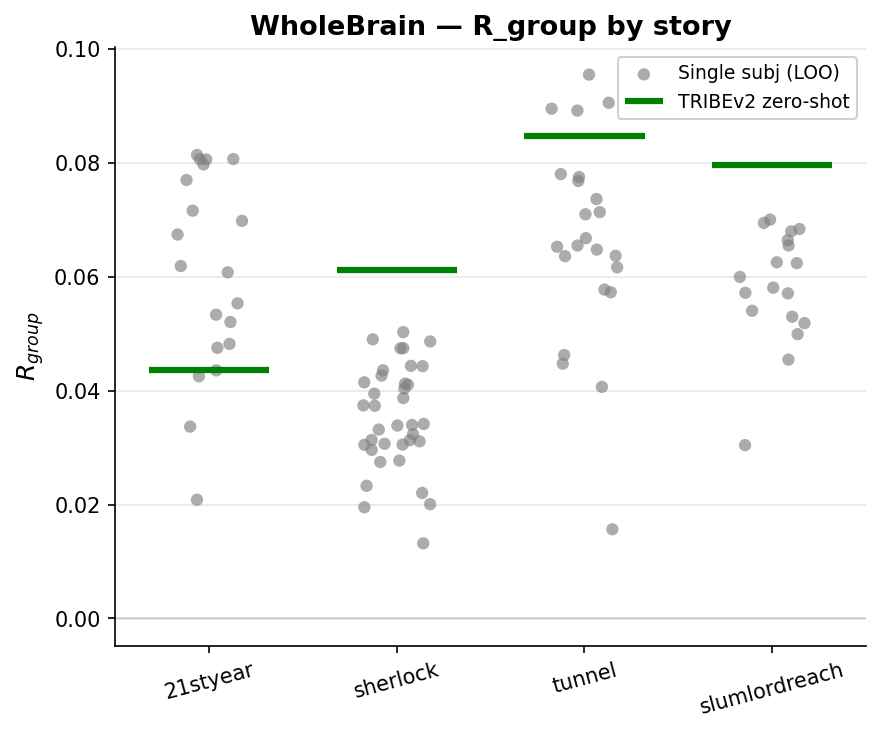}
    \end{subfigure}
    \caption{\textbf{Whole-brain $r_{\mathrm{group}}$ for TRIBEv2 across several Narratives stories.} TRIBEv2 predictions correlated against cohort-mean fMRI on the full cortical surface, with the inter-subject ceiling overlaid per story. Numbers fall within the range reported in the original TRIBEv2 paper.}
    \label{fig:tribe_validation}
\end{figure}

We run TRIBEv2 on each held-out story with the released model weights, chunking the audio at $62.4375$\,s per chunk and stitching per-chunk predictions; the stream is aligned to scanner time using the calibrated audio-onset values of App.\ref{app:eval_data:audio_onset} with the $5$\,s hemodynamic offset unmodified. Whole-brain $r_{\mathrm{group}}$ on a subset of Narratives stories falls within the range reported in the original TRIBEv2 paper (Fig.\ref{fig:tribe_validation}), confirming that our inference and alignment pipeline represents the baseline fairly.

\section{Few-shot Calibration and Baselines}
\label{app:fs_deets}

Few-shot evaluation tests whether a short calibration scan can recover listener-specific signal without refitting a full voxel-wise model.

\subsection{Calibration setup}
\label{app:fs_calib}
We use the longest single-stimulus Narratives cohort, \textit{21styear} (56 minutes per participant, $20$ participants from App.\ref{app:eval_data:dets}), as the headline benchmark; one participant (sub-$244$) is excluded for quality reasons, leaving $19$. A second, smaller-stimulus replication on \textit{slumlordreach} ($\sim$$30$ minutes, $18$ participants) is in App.\ref{app:fs_add_results}.

For each participant we hold out a fixed test segment (the last $400$ TRs of the story, $\sim$$10$ minutes) and use a contiguous prefix as the calibration window. We sweep five calibration ($C$ for short) durations: $200$, $400$, $800$, $1200$, $1600$ TRs ($\sim$$5$, $10$, $20$, $30$, $40$ minutes at TR\,=\,$1.49$\,s); the test segment is identical across calibration sizes so any difference in test-set $r$ reflects calibration size only. We report per-subject vertex-mean $r$ on the test segment, averaged across the $19$ participants with $\pm$standard-error bands. Cross-validation across the full story would let calibration size co-vary with test position, mixing the calibration-size effect with temporal-coherence effects; the fixed-test protocol avoids this confound.

\paragraph{Scope of the TRIBEv2 comparison.} We compare with TRIBEv2 in zero-shot evaluation using its released weights (App.\ref{app:tribe-baseline}). Its published fine-tuning procedure unfreezes all TRIBE parameters for one epoch (\citealp{dAscoli2026TribeV2}, Sec.~5.4). RABBiT's few-shot protocol keeps the shared encoder frozen and updates only the compact SID deviation pathway. Our few-shot comparisons use ridge and No-SID readouts fitted to the same participant-specific calibration windows; we do not report a matched few-shot comparison with TRIBEv2.

\subsection{RABBiT few-shot adaptation}
\label{app:fs_math}
RABBiT personalization freezes the backbone, the brain transformer, the SID shared bases $\Phi_i$, the shared coefficient maps $\pi_i$, and the per-subject deviation bases (initialized at the cohort average $\bar{\Delta}_i$ used at zero-shot inference). The only parameters updated are the per-ROI deviation coefficient maps $\rho_i: \mathbb{R}^{d_o} \to \mathbb{R}^{R}$, totaling $\sim$$115{,}000$ parameters across the $30$ ROIs. We minimize a per-ROI vertex-wise correlation loss with an anchor regularizer that penalizes drift of $\rho_i$ from its pretrained value:
\[
\mathcal{L}^{\mathrm{fs}} \;=\; \sum_i \mathcal{L}^{\mathrm{corr}}_i(\rho_i)\;+\; \lambda_{\rho} \sum_i \big\|\rho_i - \rho_i^{(0)}\big\|_F^2.
\]

The anchor weight $\lambda_{\rho}$ is the only material hyperparameter at calibration time. At calibration sizes $C \geq 200$ TRs, $\lambda_{\rho} = 5{\times}10^{-5}$ is at or near the held-out optimum across a $4$-subject probe sweep (per-subject optimum varies $10^{-5}$ to $10^{-3}$, but population-mean held-out $r$ is flat across this range to within $\sim$$0.005$). At smaller calibrations ($C \leq 100$) it is too weak: occasional subjects collapse below their own zero-shot performance, and a stronger $\lambda_{\rho} = 10^{-3}$ stabilizes them at the cost of $\sim$$0.005$--$0.01$ relative loss at $C \geq 800$. We use $\lambda_{\rho} = 5{\times}10^{-5}$ in all reported headline saturation curves. Optimization is AdamW for several hundred steps at a fixed learning rate, held constant across calibration sizes.

\subsection{Ridge baselines}
\label{app:fs_ridge}
We compare RABBiT few-shot calibration against two voxel-wise ridge encoders trained on the same per-subject calibration window. Both follow the canonical pipeline of \citep{huth2016natural} and \citep{antonello2024scaling}: extract per-frame audio features from a frozen backbone, build a delayed stimulus matrix using FIR taps spanning the hemodynamic window, z-score per fold, fit ridge regression per vertex via cross-validated alpha selection on a GPU implementation of bootstrap ridge.

The pretrained-features ridge uses features from a frozen Wav2Vec2.0-base (no brain-tuning) --- the canonical subject-specific encoding setup of the field. The brain-tuned-features ridge uses features from a multi-brain-tuned Wav2Vec2.0-base \citep{moussa2025multibrain} --- a backbone that has already absorbed cross-subject signal during multi-subject brain-tuning training but has no learned readout structure. The performance gap between the two ridge variants and RABBiT few-shot directly attributes RABBiT's gain to the SID readout pathway versus the backbone. A vertex-wise ridge readout on the headline $30$-ROI vertex set ($V_{\mathrm{tot}} \approx 41{,}394$) and a multi-tap FIR delay basis has $\sim$$222$M trainable parameters with the $768$-dim wav2vec backbone and $\sim$$297$M with the $1024$-dim WavLM backbone --- approximately three orders of magnitude more than the $\sim$$115{,}000$ parameters that RABBiT few-shot updates per participant.

\subsection{No-SID RABBiT (direct fine-tune) baseline}
\label{app:fs_direct}
The few-shot variant of the Per-Subject Full-Rank baseline (App.\ref{app:indp_baselines}) tests whether the gains in App.\ref{app:fs_math} come from the SID structure itself or merely from freezing most of the model and updating a small head. Architecture is identical to RABBiT except the SID readout is replaced by one per-ROI linear layer mapping $z_i$ directly to all $V_i$ vertices (no shared--deviation factorization). The full readout is $\sim$$10.6$M parameters per participant ($\sim$$92\times$ more than RABBiT few-shot's $\sim$$115{,}000$, but $\sim$$22\times$ fewer than brain-tuned ridge). At calibration time we adapt this per-ROI linear readout with the same correlation-plus-anchor objective and training schedule.

Because this baseline lacks RABBiT's strong shared-bases prior, few-shot adaptation initially \emph{loses} held-out signal at small calibrations before recovering. At $C = 200$ TRs ($\sim$$5$\,min), the No-SID readout falls to language-aggregate $r = 0.118$ versus its own zero-shot value of $r = 0.136$ --- a $\sim$$13\%$ drop. RABBiT few-shot at the same calibration size starts above its zero-shot ($r = 0.139$) and gains a further $\sim$$+0.01$ from adaptation, never going below zero-shot. The No-SID baseline only recovers parity by $C \approx 800$ TRs ($\sim$$20$\,min); this is the structural advantage SID's bounded subject-adaptation pathway provides at small calibrations.

\subsection{Additional ridge baselines}
\label{app:fs_stronger_baselines}
We evaluate two additional ridge baselines using the same few-shot calibration and held-out test protocol as in App.\ref{app:fs_calib}. Tables\ref{tab:fs_lm_tuned} and \ref{tab:fs_encoding_layer} report average percent change in prediction performance relative to zero-shot across all auditory and language ROIs, rather than only the top regions shown in Fig.\ref{fig:perroi-fewshot}. Significance is assessed against zero change using two-sided Wilcoxon signed-rank tests across participants, with one aggregate change per participant and FDR correction. The annotations $^{*}$, $^{**}$, and $^{***}$ denote FDR-corrected $p<0.05$, $p<0.01$, and $p<0.001$, respectively; $^{\mathrm{ns}}$ denotes no significant change.

\paragraph{LM-tuned backbone.}
Following prior LM-target tuning approaches \citep{vattikonda2025brainwavlm, moussa2025multibrain,moussa2025improving}, we fine-tune a speech backbone with the fMRI targets replaced by concatenated middle-layer representations of Llama2-7B for the same training stimuli. We then use its features in the same few-shot ridge setting as the pretrained and brain-tuned baselines. RABBiT few-shot yields the largest mean improvements at every calibration duration (Table\ref{tab:fs_lm_tuned}), including $+21\%$ and $+31\%$ at $5$ and $10$ minutes. Brain-tuned and LM-tuned ridge benefit from more calibration data, but their significant positive gains emerge later.

\begin{table}[htbp]
\centering
\small
\setlength{\tabcolsep}{5pt}
\renewcommand{\arraystretch}{1.15}
\begin{tabular}{@{}lrrrrr@{}}
\toprule
Model & 5\,min & 10\,min & 20\,min & 30\,min & 40\,min \\
\midrule
\textbf{RABBiT few-shot} & $\boldsymbol{+21\%}^{***}$ & $\boldsymbol{+31\%}^{***}$ & $\boldsymbol{+48\%}^{***}$ & $\boldsymbol{+52\%}^{***}$ & $\boldsymbol{+56\%}^{***}$ \\
RABBiT no-SID           & $-17\%^{***}$ & $+5\%^{\mathrm{ns}}$ & $+24\%^{***}$ & $+35\%^{***}$ & $+44\%^{***}$ \\
Ridge, brain-tuned      & $-17\%^{***}$ & $+2\%^{\mathrm{ns}}$ & $+28\%^{*}$ & $+37\%^{***}$ & $+49\%^{***}$ \\
Ridge, LM-tuned         & $-20\%^{***}$ & $-3\%^{\mathrm{ns}}$ & $+16\%^{\mathrm{ns}}$ & $+21\%^{*}$ & $+31\%^{***}$ \\
Ridge, pretrained      & $-30\%^{***}$ & $-9\%^{**}$ & $+6\%^{\mathrm{ns}}$ & $+19\%^{\mathrm{ns}}$ & $+26\%^{**}$ \\
\bottomrule
\end{tabular}
\caption{\textbf{LM-tuned ridge comparison.} Average percent change relative to zero-shot across all auditory and language ROIs. Bold values identify RABBiT few-shot. Superscripts indicate FDR-corrected significance of change from zero-shot; $\mathrm{ns}$ denotes no significant change.}
\label{tab:fs_lm_tuned}
\end{table}

\paragraph{Encoding-layer choice.}
We additionally evaluate pretrained Wav2Vec2.0-base layer $8$, motivated by prior layer-wise speech-to-brain encoding results \citep{vaidya2022self}, and compare it with layer $12$ used in the main ridge baseline. Layer $8$ improves ridge performance at all calibration durations, but RABBiT retains larger mean gains, especially with only $5$--$10$ minutes of participant-specific data (Table\ref{tab:fs_encoding_layer}). These comparisons support the advantage of RABBiT's rapid adaptation over stronger ridge alternatives in low-calibration regimes.

\begin{table}[htbp]
\centering
\small
\setlength{\tabcolsep}{5pt}
\renewcommand{\arraystretch}{1.15}
\begin{tabular}{@{}lrrrrr@{}}
\toprule
Model & 5\,min & 10\,min & 20\,min & 30\,min & 40\,min \\
\midrule
\textbf{RABBiT few-shot} & $\boldsymbol{+21\%}^{***}$ & $\boldsymbol{+31\%}^{***}$ & $\boldsymbol{+48\%}^{***}$ & $\boldsymbol{+52\%}^{***}$ & $\boldsymbol{+56\%}^{***}$ \\
Ridge, pretrained (layer 8)  & $-9\%^{*}$ & $+4\%^{\mathrm{ns}}$ & $+30\%^{***}$ & $+43\%^{***}$ & $+47\%^{***}$ \\
Ridge, pretrained (layer 12) & $-30\%^{***}$ & $-9\%^{**}$ & $+6\%^{\mathrm{ns}}$ & $+19\%^{\mathrm{ns}}$ & $+26\%^{**}$ \\
\bottomrule
\end{tabular}
\caption{\textbf{Pretrained ridge encoding-layer comparison.} Average percent change relative to zero-shot across all auditory and language ROIs. Layer $12$ is the pretrained ridge baseline reported in the main text. Bold values identify RABBiT few-shot; superscripts indicate FDR-corrected significance of change from zero-shot, with the same test and thresholds as Table\ref{tab:fs_lm_tuned}.}
\label{tab:fs_encoding_layer}
\end{table}

\subsection{Additional few-shot results}
\label{app:fs_add_results}

\begin{figure}[!t]
  \centering
  \includegraphics[width=4.914in]{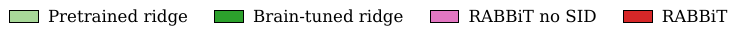}\\[6pt]
  \includegraphics[width=0.69\linewidth]{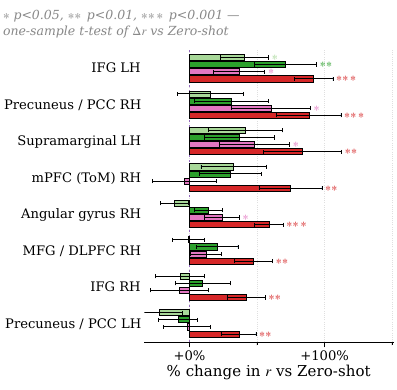}
  \caption{\textbf{Regional gains after 10 minutes of few-shot calibration.} Percent change in prediction correlation relative to zero-shot for the top ROIs on \textit{21styear}, comparing RABBiT few-shot, No-SID direct fine-tuning, brain-tuned ridge, and pretrained ridge at $C=400$ TRs ($\sim$10 minutes). RABBiT's largest gains occur in higher-order language regions ($30$--$90\%$ relative improvement); results for other calibration sizes follow in Fig.\ref{fig:perroi_21st_calibsizes}.}
  \label{fig:perroi-fewshot}
\end{figure}
\begin{figure}[!htb]
    \centering
        \includegraphics[width=4.914in]{figs/legend.pdf}

    \begin{subfigure}[b]{0.8\textwidth}
        \includegraphics[width=\linewidth]{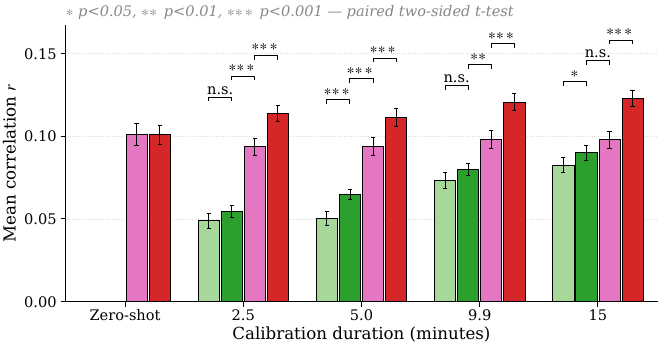}
    \end{subfigure}
    \caption{\textbf{Few-shot saturation curve on \textit{slumlordreach}.} Per-subject mean correlation $r$ on the held-out test segment versus calibration duration on $18$ participants}
    \label{fig:slumlord_saturation}
\end{figure}

Table\ref{tab:fewshot_stats} reports paired two-sided $t$-tests across the $19$-subject panel between consecutive sorted conditions of the saturation curve. RABBiT few-shot significantly outperforms the No-SID direct fine-tune at every calibration size from $5$ to $40$ minutes ($p < 0.001$ for $5$--$30$\,min, $p \approx 0.006$ at $40$\,min). The No-SID direct fine-tune does not significantly outperform the brain-tuned ridge baseline from $10$ minutes onward (numerical $\Delta r \approx +0.003$--$+0.005$, $p > 0.10$).

\begin{table}[h]
\centering
\small
\begin{tabular}{lrrr}
\toprule
Calibration & ridge $\to$ direct & direct $\to$ RABBiT few-shot & ridge $\to$ RABBiT few-shot \\
\midrule
 5\,min ($C{=}200$)  & $\Delta\,{+}0.009$, $p{=}0.029$\,*\,    & $\Delta\,{+}0.030$, $p{<}0.001$\,***\, & ***  \\
10\,min ($C{=}400$)  & $\Delta\,{+}0.005$, $p{=}0.108$\,n.s.\, & $\Delta\,{+}0.021$, $p{<}0.001$\,***\, & ***  \\
20\,min ($C{=}800$)  & $\Delta\,{+}0.003$, $p{=}0.201$\,n.s.\, & $\Delta\,{+}0.014$, $p{<}0.001$\,***\, & ***  \\
30\,min ($C{=}1200$) & $\Delta\,{+}0.004$, $p{=}0.112$\,n.s.\, & $\Delta\,{+}0.009$, $p{=}0.0001$\,***\, & ***  \\
40\,min ($C{=}1600$) & $\Delta\,{+}0.003$, $p{=}0.172$\,n.s.\, & $\Delta\,{+}0.005$, $p{=}0.006$\,**\,   & **   \\
\bottomrule
\end{tabular}
\caption{\textbf{Pairwise paired two-sided $t$-tests on the language-aggregate saturation curve.} $n{=}19$ participants on \textit{21styear}; $\Delta r$ between consecutive sorted conditions.}
\label{tab:fewshot_stats}
\end{table}

\begin{figure}[h]
    \centering
    \includegraphics[width=4.914in]{figs/legend.pdf}
    \begin{subfigure}[t]{0.69\textwidth}
        \includegraphics[width=\linewidth]{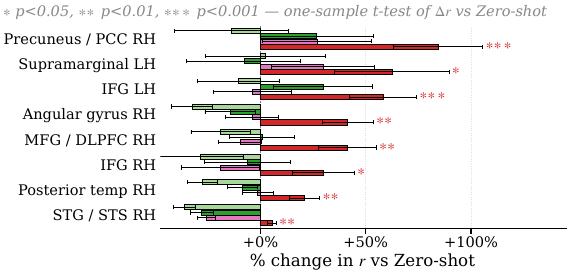}
        \caption{$C = 200$ TRs ($\sim$$5$\,min)}
        \label{fig:perroi_21st_C200}
    \end{subfigure}\hfill
    \begin{subfigure}[t]{0.69\textwidth}
        \includegraphics[width=\linewidth]{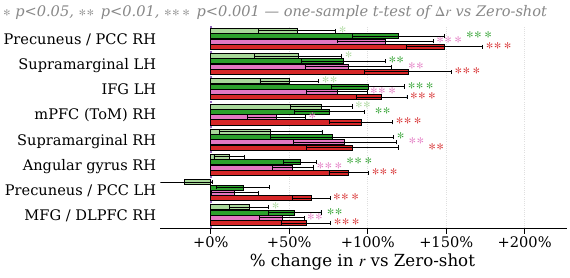}
        \caption{$C = 800$ TRs ($\sim$$20$\,min)}
        \label{fig:perroi_21st_C800}
    \end{subfigure}

    \vspace{4pt}
    \begin{subfigure}[t]{0.69\textwidth}
        \includegraphics[width=\linewidth]{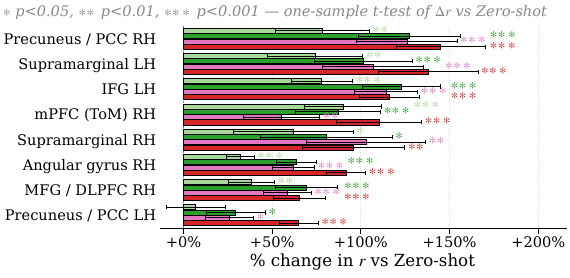}
        \caption{$C = 1200$ TRs ($\sim$$30$\,min)}
        \label{fig:perroi_21st_C1200}
    \end{subfigure}\hfill
    \begin{subfigure}[t]{0.69\textwidth}
        \includegraphics[width=\linewidth]{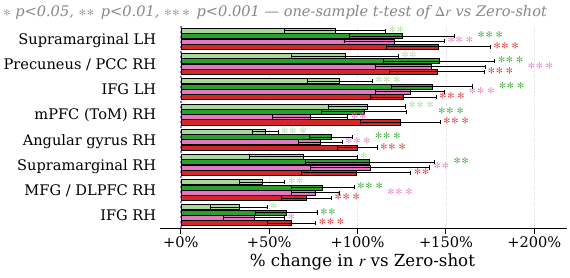}
        \caption{$C = 1600$ TRs ($\sim$$40$\,min)}
        \label{fig:perroi_21st_C1600}
    \end{subfigure}
    \caption{\textbf{Per-ROI percent change in $r$ relative to zero-shot on \textit{21styear}, across four additional calibration sizes.} Each panel reports per-ROI deltas for RABBiT few-shot, brain-tuned ridge, and pretrained-features ridge. $C{=}400$ panel is shown in Fig.\ref{fig:perroi-fewshot}.}
    \label{fig:perroi_21st_calibsizes}
\end{figure}

\begin{figure}[h]
    \centering
    \includegraphics[width=4.914in]{figs/legend.pdf}
    \begin{subfigure}[t]{0.69\textwidth}
        \includegraphics[width=\linewidth]{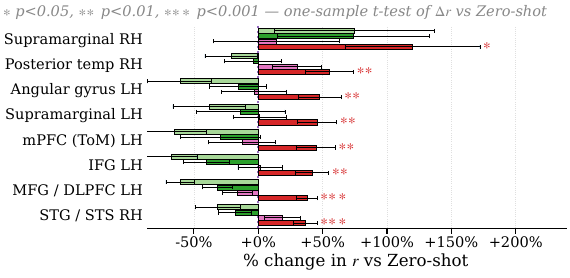}
        \caption{$C = 100$ TRs ($\sim$$2.5$\,min)}
        \label{fig:perroi_slum_C100}
    \end{subfigure}\hfill
    \begin{subfigure}[t]{0.69\textwidth}
        \includegraphics[width=\linewidth]{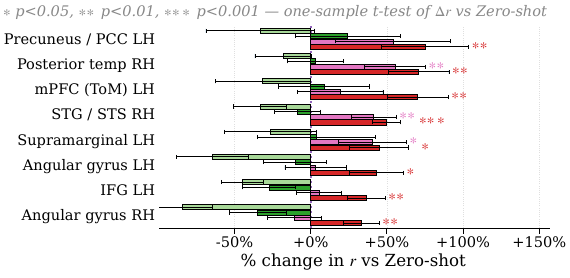}
        \caption{$C = 200$ TRs ($\sim$$5$\,min)}
        \label{fig:perroi_slum_C200}
    \end{subfigure}

    \vspace{4pt}
    \begin{subfigure}[t]{0.69\textwidth}
        \includegraphics[width=\linewidth]{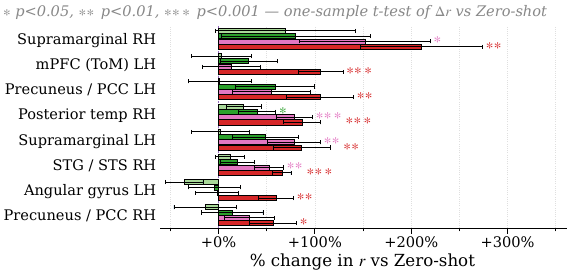}
        \caption{$C = 400$ TRs ($\sim$$10$\,min)}
        \label{fig:perroi_slum_C400}
    \end{subfigure}\hfill
    \begin{subfigure}[t]{0.69\textwidth}
        \includegraphics[width=\linewidth]{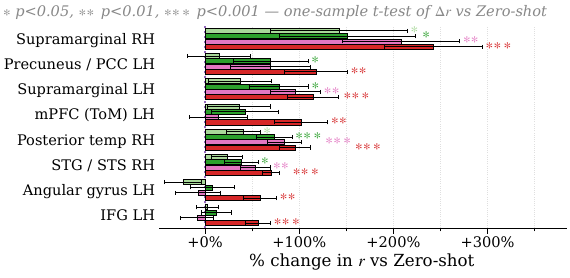}
        \caption{$C = 600$ TRs ($\sim$$15$\,min)}
        \label{fig:perroi_slum_C600}
    \end{subfigure}
    \caption{\textbf{Per-ROI percent change in $r$ relative to zero-shot on \textit{slumlordreach}, across four calibration sizes.} Each panel includes the No-SID direct fine-tune as a comparator.}
    \label{fig:perroi_slum_calibsizes}
\end{figure}

The per-ROI comparison in Fig.\ref{fig:perroi-fewshot} shows the per-ROI percent change at $C{=}400$. Fig.\ref{fig:perroi_21st_calibsizes} gives the per-ROI breakdown at the four other calibration sizes ($C \in \{200, 800, 1200, 1600\}$) on \textit{21styear}: the qualitative ROI ordering is preserved, the magnitude grows monotonically with calibration size, and the largest improvements remain concentrated in higher-order language and DMN ROIs (IFG, supramarginal, mPFC, angular gyrus, MFG/DLPFC, precuneus/PCC). Fig.\ref{fig:slumlord_saturation} and Fig.\ref{fig:perroi_slum_calibsizes} give the corresponding saturation curve and per-ROI breakdown on \textit{slumlordreach} ($18$ participants, calibration sizes $\{100, 200, 400, 600\}$ TRs); at $C{=}400$ on the wav2vec backbone, RABBiT few-shot reaches $r = 0.114$ on the LH-language aggregate vs $r = 0.098$ for No-SID direct and $r = 0.094$ for brain-tuned ridge. Two methodological notes: \textit{slumlordreach} has a different optimal residual hemodynamic shift than \textit{21styear} ($\Delta t \approx 2$ vs $3$ TRs), and we use per-story-calibrated comparisons throughout; the headline anchor $\lambda_{\rho} = 5{\times}10^{-5}$ used in all reported saturation curves is appropriate for $C \geq 200$ TRs (a stronger $\lambda_{\rho} = 10^{-3}$ is recommended for very-small-calibration regimes $C \leq 100$).

\begin{figure}[!htbp]
    \centering

    \begin{subfigure}[t]{0.69\textwidth}
        \includegraphics[width=\linewidth]{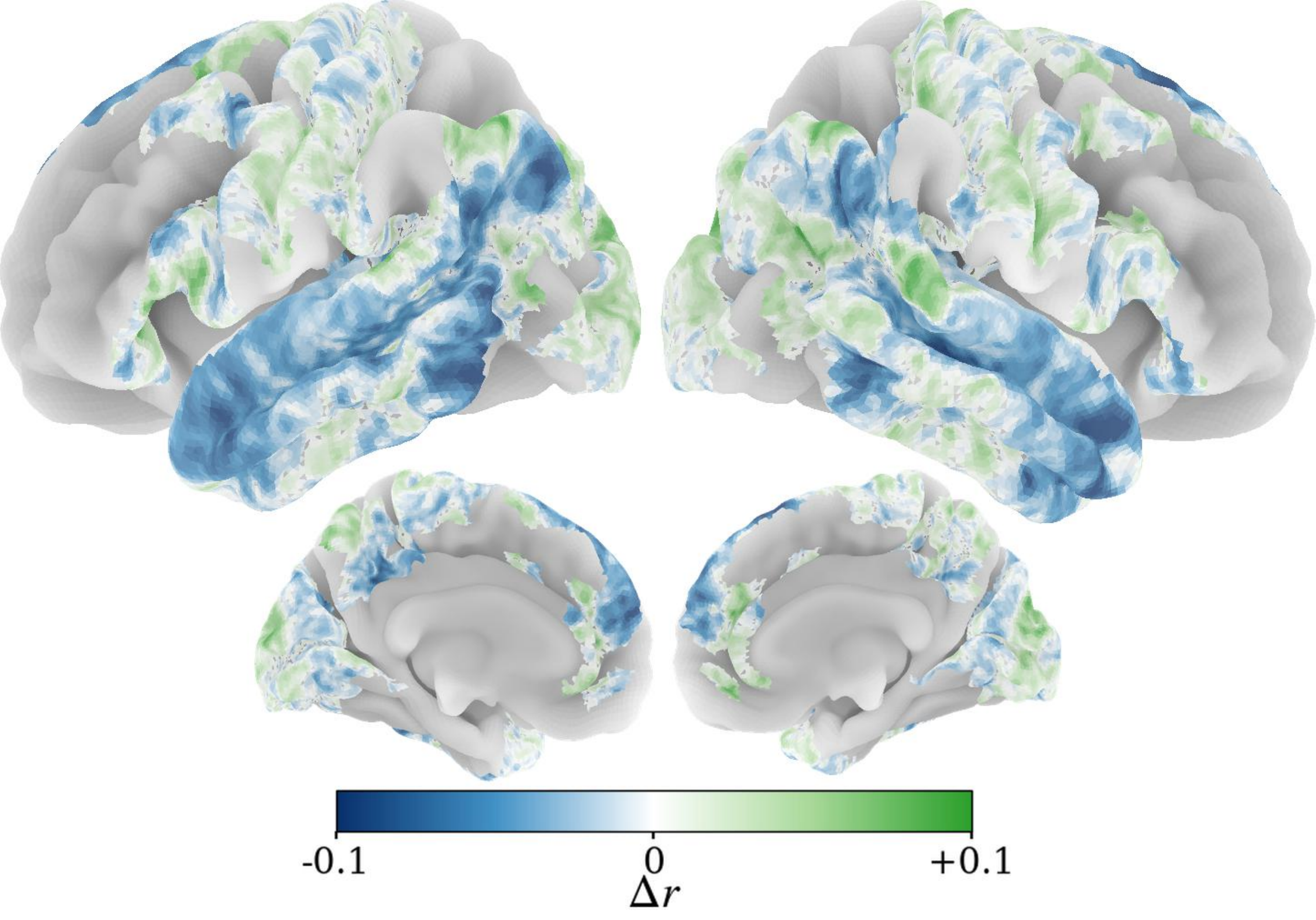}
        \caption{Pretrained ridge --- zero-shot for $C = 400$ TRs ($\sim10$\,min)}
        \label{fig:brain_ridge_fs}
    \end{subfigure}\\[4pt]

    \begin{subfigure}[t]{0.69\textwidth}
        \includegraphics[width=\linewidth]{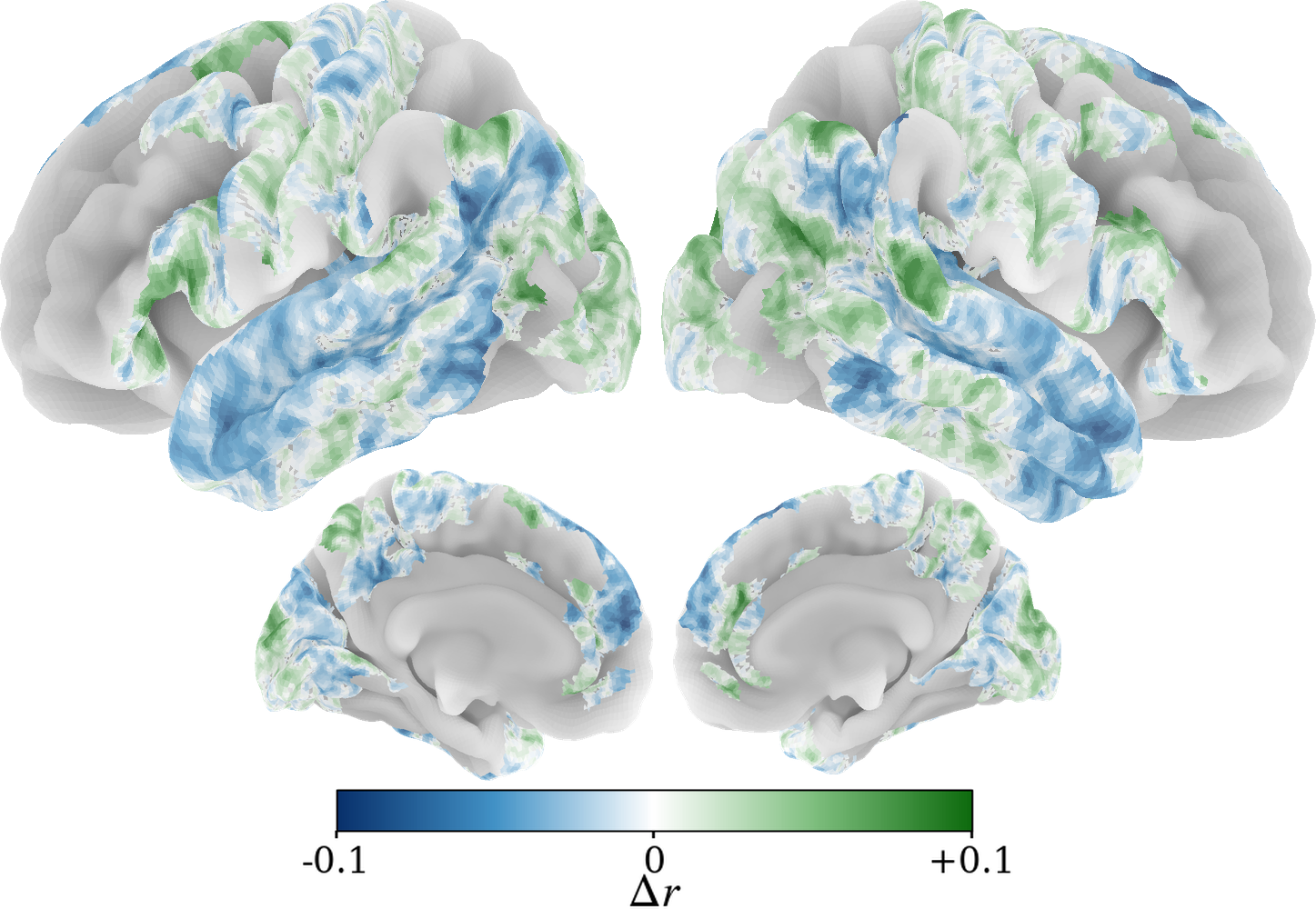}
        \caption{Brain-tuned ridge --- zero-shot for $C = 400$ TRs ($\sim10$\,min)}
        \label{fig:brain_ridge_bt_fs}
    \end{subfigure}\\[4pt]

    \begin{subfigure}[t]{0.69\textwidth}
        \includegraphics[width=\linewidth]{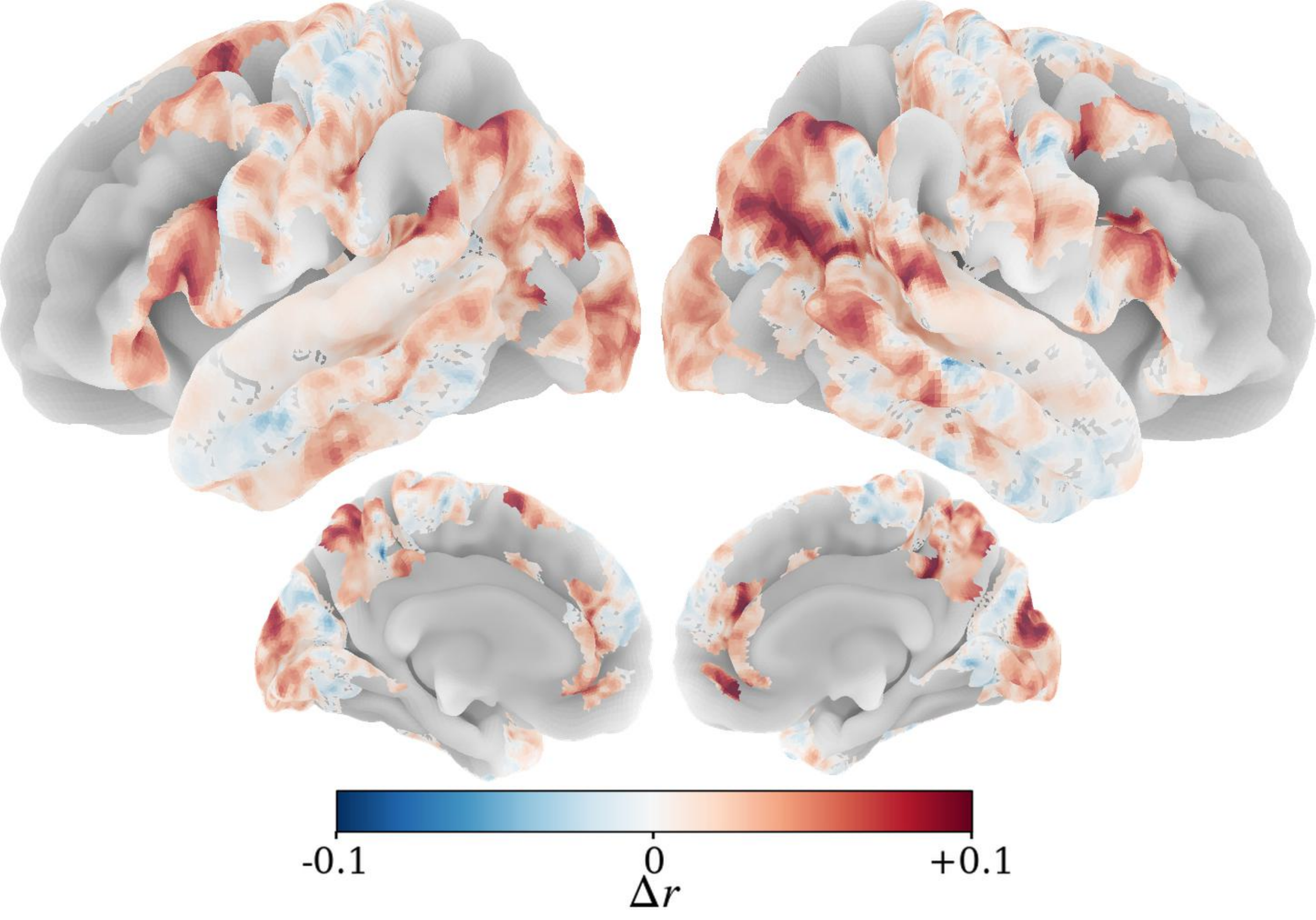}
        \caption{Few-shot --- zero-shot for $C = 800$ TRs ($\sim20$\,min)}
        \label{fig:brain_fs_800}
    \end{subfigure}

    \caption{\textbf{Additional brain maps: change in $r$ relative to zero-shot on \textit{21styear}.}}
    \label{fig:brain_21st_calibsizes}
\end{figure}

\section{Spatial autocorrelation and ROI granularity}
\label{app:spatial_controls}
Exploiting spatial autocorrelation can inflate performance results. Spatially correlated fMRI signals can make it difficult to distinguish prediction of a shared regional response from prediction of finer spatial structure. We therefore examine the statistical units used for inference, compare models after averaging or removing ROI-wide activity, and measure RABBiT's predictive contribution beyond ridge on a separate held-out segment. These analyses address complementary aspects of spatial dependence, but they don't conclude the risk is eliminated entirely. 

\subsection{Statistical units and spatial dependence}
\label{app:spatial_inference}
The main zero-shot model comparisons use per-story or per-section scores, with one scalar per story or section for the region being contrasted. Few-shot comparisons use participants as the sampling units. Thus, these encoding-performance comparisons do not treat cortical vertices as independent observations. Query-assignment controls separately assess embedding permutations and consistency across ROIs (App.\ref{app:query_permutation}). The language-localizer analysis instead tests the contrast across $32$ matched intact/degraded stimulus pairs at each vertex, with FDR correction (Sec.\ref{sec:methods-ablation}); its observations are stimulus pairs. Spatial dependence nevertheless remains relevant to the interpretation of prediction scores, and a large vertex count does not substitute for additional independent stories or participants.

\subsection{ROI-mean and within-ROI residual prediction}
\label{app:spatial_residuals}
We compare RABBiT few-shot with voxel-wise ridge using pretrained Wav2Vec2.0 layer-$12$ features on \textit{21styear} ($20$ participants), using matched calibration and held-out segments for both models. For ROI $i$ with $V_i$ vertices, we first average the measured and predicted signals separately:
\[
\bar y_i(t)=\frac{1}{V_i}\sum_{v\in i}y_{i,v}(t),
\qquad
\overline{\hat y}_i(t)=\frac{1}{V_i}\sum_{v\in i}\hat y_{i,v}(t).
\]
The ROI-mean score is $r_{\mathrm{mean}}=\mathrm{corr}_t(\bar y_i,\overline{\hat y}_i)$, computed after spatial averaging. This tests whether the performance advantage persists when each ROI contributes a single timecourse. To test prediction beyond that common regional component, we subtract each signal's own ROI mean from every vertex:
\[
y^{\mathrm{resid}}_{i,v}(t)=y_{i,v}(t)-\bar y_i(t),
\qquad
\hat y^{\mathrm{resid}}_{i,v}(t)=\hat y_{i,v}(t)-\overline{\hat y}_i(t),
\]
and evaluate prediction correlations on the resulting within-ROI residuals. We compare one aggregate score per participant using paired, two-sided Wilcoxon signed-rank tests across participants, with FDR correction. Table\ref{tab:spatial_correlations} reports both evaluations averaged across all language regions, including all five calibration durations.

\begin{table}[htbp]
\centering
\small
\setlength{\tabcolsep}{4pt}
\renewcommand{\arraystretch}{1.15}
\begin{tabular}{@{}lrrrrrr@{}}
\toprule
& \multicolumn{3}{c}{ROI-mean timecourse ($r_{\mathrm{mean}}$)} & \multicolumn{3}{c}{Within-ROI residual ($r_{\mathrm{resid}}$)} \\
\cmidrule(lr){2-4}\cmidrule(l){5-7}
Calibration & RABBiT & Ridge & $p_{\mathrm{FDR}}$ & RABBiT & Ridge & $p_{\mathrm{FDR}}$ \\
\midrule
5\,min  & \textbf{0.1764} & 0.0676 & $1.9\times10^{-6}$ & \textbf{0.0686} & 0.0252 & $1.9\times10^{-6}$ \\
10\,min & \textbf{0.1790} & 0.0871 & $1.9\times10^{-6}$ & \textbf{0.0731} & 0.0328 & $1.9\times10^{-6}$ \\
20\,min & \textbf{0.1862} & 0.1163 & $1.9\times10^{-6}$ & \textbf{0.0766} & 0.0471 & $1.9\times10^{-6}$ \\
30\,min & \textbf{0.1930} & 0.1341 & $1.9\times10^{-6}$ & \textbf{0.0790} & 0.0581 & $1.9\times10^{-6}$ \\
40\,min & \textbf{0.2043} & 0.1520 & $8.2\times10^{-5}$ & \textbf{0.0802} & 0.0693 & $1.1\times10^{-2}$ \\
\bottomrule
\end{tabular}
\caption{\textbf{Prediction after averaging or removing ROI-wide activity.} Correlations averaged across language regions and participants; bold values indicate the larger model score. The $p$-values compare RABBiT few-shot with pretrained layer-$12$ ridge using paired, two-sided Wilcoxon signed-rank tests across $20$ participants, with FDR correction.}
\label{tab:spatial_correlations}
\end{table}

RABBiT has higher correlations in both evaluations at every calibration duration. Ridge improves with more data, and the residual gap narrows at $40$ minutes, but the advantage remains largest in the low-calibration regime. Persistence after removing the ROI mean indicates that shared regional activity alone does not explain the gains. Residual signals can still be spatially correlated, so this analysis does not remove all forms of spatial dependence.

\subsection{Unique predictive variance beyond ridge}
\label{app:spatial_unique_variance}
To assess whether RABBiT captures signal complementary to ridge, we partition \textit{21styear} ($20$ participants) into three disjoint temporal segments: $A$ ($24.8$ minutes), $B$ ($17.5$ minutes), and $C$ ($13$ minutes). We leave $10$-TR gaps between adjacent segments to reduce temporal-autocorrelation leakage. This analysis uses a separate three-way split from the calibration-duration sweep above.

On $A$, we fit a ridge encoder using pretrained Wav2Vec2.0 layer-$12$ representations and perform RABBiT few-shot adaptation. On $B$, we regress measured responses on both models' predictions using banded ridge, and separately fit ridge calibrators using each model's predictions alone. We evaluate all three combinations on $C$, unused for either fitting stage, to obtain $R^2_{\mathrm{joint}}$, $R^2_{\mathrm{ridge}}$, and $R^2_{\mathrm{RABBiT}}$.

We estimate RABBiT's unique contribution as the increment in held-out explained variance when its predictions are added to ridge:
\[
U_{\mathrm{RABBiT}}=R^2_{\mathrm{joint}}-R^2_{\mathrm{ridge}}.
\]
For each evaluation, the squared leave-one-out inter-subject correlation on $C$ provides an approximate explainable-variance estimate, $R^2_{\mathrm{ceiling}}\approx\mathrm{ISC}_{\mathrm{LOO}}^2$. We express each score as $100\,R^2/R^2_{\mathrm{ceiling}}$ and the unique contribution as $100\,U_{\mathrm{RABBiT}}/R^2_{\mathrm{ceiling}}$. The estimated ceilings are $0.095$ for ROI means and $0.0215$ for within-ROI residuals.

\begin{table}[htbp]
\centering
\small
\setlength{\tabcolsep}{5pt}
\renewcommand{\arraystretch}{1.15}
\begin{tabular}{@{}lrrrr@{}}
\toprule
& \multicolumn{2}{c}{ROI-mean timecourse} & \multicolumn{2}{c}{Within-ROI residual} \\
\cmidrule(lr){2-3}\cmidrule(l){4-5}
Model or contribution & $R^2$ & \% ceiling & $R^2$ & \% ceiling \\
\midrule
Ridge              & 0.0273 & 28.7   & 0.0055 & 25.6 \\
RABBiT             & 0.0479 & 50.4   & 0.0101 & 47.0 \\
Joint              & 0.0483 & 50.8   & 0.0099 & 46.046 \\
\textbf{Unique RABBiT} & \textbf{0.0210} & \textbf{22.1} & \textbf{0.0044} & \textbf{20.46} \\
\bottomrule
\end{tabular}
\caption{\textbf{Held-out variance partitioning on \textit{21styear}.} Scores summarize the language regions on segment $C$, after base-model fitting on $A$ and prediction-combination fitting on $B$. Percentages are relative to the approximate explainable-variance estimates, $R^2_{\mathrm{ceiling}}\approx0.095$ for ROI means and $0.0215$ for residuals. The unique row reports $U_{\mathrm{RABBiT}}$, the joint-minus-ridge increment in $R^2$.}
\label{tab:spatial_unique_variance}
\end{table}

RABBiT's unique increments correspond to approximately $22\%$ of estimated explainable variance for ROI means and $20\%$ for residuals (Table\ref{tab:spatial_unique_variance}). The joint residual score is slightly below RABBiT alone; regularized combinations are not guaranteed to improve held-out $R^2$. The positive joint-minus-ridge increments support a predictive contribution beyond ridge under this split.

\subsection{Functional interpretation and ROI granularity}
\label{app:spatial_interpretation}
RABBiT's intact-minus-degraded language-localizer contrast recovers the canonical left-lateralized network without training on the controlled localizer (Sec.\ref{sec:methods-ablation}; Fig.\ref{fig:fed-ln}). This provides complementary evidence of functional selectivity, while the controls above show predictive gains beyond ROI-wide cofluctuations.

The ROI bottleneck controls temporal routing, specifying the speech context available to each region. SID readouts construct separate predictions for individual vertices, without requiring a homogeneous code within each ROI (Sec.\ref{sec:methods-shareddev-form}). The formulation accommodates finer or coarser parcellations, grouped parcels, or expanded coverage up to the whole cortex (App.\ref{app:whole_cortex}). ROI granularity is a modeling choice for routing and parameter efficiency; neither that choice nor higher encoding accuracy identifies a mechanistic neural code. 

\clearpage

\end{document}